\documentclass[11pt]{article}

\usepackage{setspace,graphicx,epstopdf,amsmath,amsfonts,amssymb,amsthm,versionPO}
\usepackage{marginnote,datetime,enumitem,rotating,fancyvrb,caption,subcaption}
\usepackage{hyperref,float}

\usepackage{natbib}
\usepackage{titlesec} %required for subsubsubsection
\usdate

\excludeversion{notes}		% Include notes?
\includeversion{links}          % Turn hyperlinks on?

\iflinks{}{\hypersetup{draft=true}}

\ifnotes{%
\usepackage[margin=1in,paperwidth=10in,right=2.5in]{geometry}%
\usepackage[textwidth=1.4in,shadow,colorinlistoftodos]{todonotes}%
}{%
\usepackage[margin=1in]{geometry}%
\usepackage[disable]{todonotes}%
}

\makeatletter\let\chapter\@undefined\makeatother % Undefine \chapter for todonotes

\usepackage{indentfirst} % Indent first sentence of a new section.
\usepackage{endnotes}    % Use endnotes instead of footnotes
\usepackage{jf}          % JF-specific formatting of sections, etc.
\usepackage{wrapfig}
\usepackage[labelfont=bf,labelsep=period]{caption}   % Format figure captions
\newtheorem{theorem}{THEOREM}
\newtheorem{lemma}{LEMMA}
\newtheorem{refproof}{PROOF}

\usepackage{amsmath,amsfonts,bm}

\def\eqref#1{equation~\ref{#1}}
\def\1{\bm{1}}

\def\rxi{{\bm{\xi}}}

\def\ra{{\bm{a}}}

\def\rh{{\bm{h}}}

\def\rx{{\bm{x}}}
\def\ry{{\bm{y}}}
\def\rz{{\bm{z}}}

\DeclareMathAlphabet{\mathsfit}{\encodingdefault}{\sfdefault}{m}{sl}
\SetMathAlphabet{\mathsfit}{bold}{\encodingdefault}{\sfdefault}{bx}{n}

\newcommand{\E}{\mathbb{E}}

\makeatletter
\def\blfootnote{\gdef\@thefnmark{}\@footnotetext}
\makeatother

\begin{document}

\setlist{noitemsep}  % Reduce space between list items (itemize, enumerate, etc.)
\onehalfspacing      % Use 1.5 spacing
% Use endnotes instead of footnotes - redefine \footnote command
\renewcommand{\footnote}{\endnote}  % Endnotes instead of footnotes

\author{Jonas Rothfuss \thanks{denotes equal contribution} \thanks{The authors are with the Computational Risk and Asset Management Research Group at the Karlsruhe Institute of Technology (KIT), Germany. Reference to \texttt{jonas.rothfuss@gmail.com}}. 
\and Fabio Ferreira\footnotemark[1] \footnotemark[2]  \and
Simon Walther\footnotemark[2] \and
Maxim Ulrich\footnotemark[2]
}

\title{\Large \bf Conditional Density Estimation with Neural Networks: Best Practices and Benchmarks}
\date{}              % No date for final submission

% Create title page with no page number
\maketitle
\thispagestyle{empty}

\bigskip

\centerline{\bf ABSTRACT}

\begin{doublespace}  % Double-space the abstract and don't indent it
  \noindent % Given a set of empirical observations, conditional density estimation aims to capture the statistical relationship between a conditional variable $\rx$ an dependent variable $\ry$ by modelling their conditional probability $p(\ry|\rx)$. This is important for many tasks in econometrics and finance that are not only interested in the conditional mean, but also involve modelling deviations from the mean.

Given a set of empirical observations, conditional density estimation aims to capture the statistical relationship between a conditional variable $\mathbf{x}$ and a dependent variable $\mathbf{y}$ by modeling their conditional probability $p(\mathbf{y}|\mathbf{x})$. 
%
% This important for many tasks in econometrics and finance that are not only interested in the conditional mean, but also aim at modelling deviations from the mean and the associated risk.
%
The paper develops best practices for conditional density estimation for finance applications with neural networks, grounded on mathematical insights and empirical evaluations. In particular, we introduce a noise regularization and data normalization scheme, alleviating problems with over-fitting, initialization and hyper-parameter sensitivity of such estimators. We compare our proposed methodology with popular semi- and non-parametric density estimators, underpin its effectiveness in various benchmarks on simulated and Euro Stoxx 50 data and show its superior performance. Our methodology allows to obtain high-quality estimators for statistical expectations of higher moments, quantiles and non-linear return transformations, with very little assumptions about the return dynamic.
\end{doublespace}

\medskip

%\noindent JEL classification: XXX, YYY.

\clearpage

% \noindent Note that the JF doesn't want the first section to be titled, and the text here is not indented.\footnote{Here's a sample footnote (endnote).} Let's put in some sections and subsections to see how they get formatted.

% \clearpage

%%%%%%%%%%%%%%%%%%%% Main content inputs %%%%%%%%%%%%%%%%%%%%%%%%%%%%%%%%%

\section{Introduction}
A wide range of problems in econometrics and finance are concerned with describing the statistical relationship between a vector of explanatory variables $\rx$ and a dependent variable or vector $\ry$ of interest. While regression analysis aims to describe the conditional mean $\mathbb{E}[\ry|\rx]$, many problems in risk and asset management require gaining insight about deviations from the mean and their associated likelihood. The stochastic dependency of $\ry$ on $\rx$ can be fully described by modeling the conditional probability density $p(\ry|\rx)$. Inferring such a density function from a set of empirical observations $\{(\rx_n, \ry_n)\}^N_{n=1}$ is typically referred to as conditional density estimation (CDE).

We propose to use neural networks for estimating conditional densities. In particular, we discuss two models in which a neural network controls the parameters of a Gaussian mixture. Namely, these are the Mixture Density Network (MDN) by \citet{Bishop1994} and the Kernel Mixture Network (KMN) by \citet{Ambrogioni2017}. When chosen expressive enough, such models can approximate arbitrary conditional densities. 

However, when combined with maximum likelihood estimation, this flexibility can result in over-fitting and poor generalization beyond the training data. Addressing this issue, we develop a noise regularization method for conditional density estimation. By adding small random perturbations to the data during training, the conditional density estimate is smoothed and generalizes better. In fact, we mathematically derive that adding noise during training is equivalent to penalizing the second derivatives of the conditional log-probability. Graphically, the penalization punishes very curved or even spiky density estimators in favor of smoother variants. Our experimental results demonstrate the efficacy and importance of the noise regularization for attaining good out-of-sample performance.

Moreover, we attend to further practical issues that arise due to different value ranges of the training data. In this context, we introduce a simple data normalization scheme, that fits the conditional density model on normalized data, and, after training, transforms the density estimate, so that is corresponds to the original data distribution. The normalization scheme makes the hyper-parameters and initialization of the neural network based density estimator insensitive to differing value ranges. Our empirical evaluations suggest that this increases the consistency of the training results and significantly improves the estimator's performance.

Aiming to compare our proposed approach against well-established CDE methods, we report a comprehensive benchmark study on simulated densities as well as on EuroStoxx 50 returns. When trained with noise regularization, both MDNs and KMNs are able to outperform previous standard semi- and nonparametric conditional density estimators. Moreover, the results suggest that even for small sample sizes, neural network based conditional density estimators can be an equal or superior alternative to well established conditional kernel density estimators. 

Our study adds to the econometric literature, which discusses two main approaches towards CDE. The majority of financial research assumes that the conditional distribution follows a standard parametric family (e.g. Gaussian) that captures the dependence of the distribution parameters on $\rx$ with a (partially) linear model. The widely used ARMA-GARCH time-series model \citep{Engle1982, Nelson1992} and many of its extensions \citep{Glosten1993, Hansen1994, Sentana1995} fall into this category. However, inherent assumptions in many such models have been refuted empirically later on \citep{Harvey1999, Jondeau2003}. Another example for this class of models are linear factor models \citep{Fama1993, Carhart1997, Fama2015}. Here, too, evidence for time variation in the betas of these factor models, as documented by \citet{Jagannathan1996}, \citet{Lewellen2006} or \citet{Gormsen2017}, cast doubt about the actual existence of the stated linear relationships. Overall, it is unclear to which degree the modelling restrictions are consistent with the actual mechanisms that generate the empirical data and how much they bias the inference. 

Another major strand of research approaches CDE from a nonparametric perspective, estimating the conditional density with kernel functions, centered in the data points \citep{Hyndman1996, Li2007}.
% Maybe you can find 1 or 2 papers from finance that use this approach?
While kernel methods make little assumptions about functional relationships and density shape, they typically suffer from poor generalization in the tail regions and from data sparseness when dimensionality is high.

In contrast, CDE based on high-capacity function approximators such as neural networks has received little attention in the econometric and finance community. Yet, they combine the global generalization capabilities of parametric models with little restrictive assumptions regarding the conditional density. Aiming to combine these two advantages, this work studies the use of neural networks for estimating conditional densities. 
Overall, this paper establishes a sound framework for fitting high-capacity conditional density models. Thanks to the presented noise regularization and data normalization scheme, we are able to overcome common issues with neural network based estimators and make the approach easy to use. The conditional density estimators are available as open-source python package.\footnotemark[1]

\footnotetext[1]{\href{https://github.com/freelunchtheorem/Conditional_Density_Estimation}{https://github.com/freelunchtheorem/Conditional\_Density\_Estimation}}

\section{Background}
\label{sec:background}
%%%%%%%%%%%%%%%%%%%%%%%%%%%%%%%%%%%%%%%%%%%%%%%%%%%%%%%%%%%%%%%%%%%%%%%%%%%%%%%%%%
\subsection{Density Estimation}
Let $X$ be a random variable with probability density function (PDF) $p(x)$ defined over the domain  $\mathcal{X}$. When investigating phenomena in the real world, the distribution of an observable variable $X$ is typically unknown. However, it possible to observe realizations $\rx_{n} \sim p(\rx)$ of $X$. Given a collection $\mathcal{D} = \{\rx_{1}, ..., \rx_{n} \}$ of such observations, it is our aim to find a good estimate $\hat{p}(x)$ of the true density function $p$.
Typically, the goodness of a fitted distribution $\hat{p}$ is measured by a statistical divergence between the estimated $\hat{p}$ and the true density function $p$. Throughout the density estimation literature \citep{Bishop2006, Li2007, Shalizi2011} the most common criterions are integrated mean squared error (IMSE) and the Kullback-Leibler divergence. 

In its most general form, density estimation aims to find the best $\hat{p}$ among all possible PDFs over the domain $\mathcal{X}$, while only given a finite number of observations. Even in the simple case $\mathcal{X}=\mathbb{R}^1$, this would require estimating infinitely many distribution parameters with a finite amount of data, which is not feasible in practice. 
Hence, it is necessary to either restrict the space of possible PDFs or to embed other assumptions into the density estimation. The kind of imposed assumptions characterize the distinction between the sub-fields of parametric and non-parametric density estimation.

%%%%%%%%%%%%%%%%%%%%%%%%%%%%%%%%%%%%%%%%%%%%%%%%%%%%%%%%%%%%%%%%%%%%%%%%%%%%%%%%%%%
\subsubsection{Parametric Density Estimation}
In parametric estimation, the PDF $\hat{p}$ is assumed to belong to a parametric family $\mathcal{F} = \{ \hat{p}_\theta(\cdot) | \theta \in \Theta \}$
where the density function is described by a finite dimensional parameter $\theta \in \Theta$. A classical example of $\mathcal{F}$ is the family of univariate normal distributions $\{ \mathcal{N}(~\cdot~ |\mu, \sigma) | (\mu,\sigma) \in \mathbb{R} \times \mathbb{R}^+ \}$.

The standard method for estimating $\theta$ is \textit{maximum likelihood estimation}, wherein $\theta^*$ is chosen so that the likelihood of the data $\mathcal{D}$ is maximized:
\begin{equation}
\theta^* = \text{arg} \max_{\theta} \prod_{n=1}^N  \hat{p}_\theta(\rx_{n}) =  \text{arg} \max_{\theta} \sum_{n=1}^N  \log \hat{p}_\theta(\rx_{n})
\label{eq:mle-to-kl1}
\end{equation}
In practice, the optimization problem is restated as maximizing the sum of log-probabilities which is equivalent to minimizing the Kullback-Leibler divergence between the empirical data distribution $ p_\mathcal{D}(x) = \frac{1}{N} \sum_{n=1}^N \delta(||\rx-\rx_{n}||)$ (i.e. point mass in the observations $\rx_{n}$) and the parametric distribution $\hat{p}_\theta$:
\begin{align}
    \theta^* = \text{arg} \min_{\theta} \mathcal{D}_{KL}(p_\mathcal{D}||\hat{p}_\theta) \label{eq:mle-to-kl2}
\end{align}
For further details on parametric density estimation, we refer to \citet{Bishop2006} page 57.

%%%%%%%%%%%%%%%%%%%%%%%%%%%%%%%%%%%%%%%%%%%%%%%%%%%%%%%%%%%%%%%%%%%%%%%%%%%%%%%%%%%%%
\subsubsection{Nonparametric Density Estimation} \label{sec:nonparametric}
In contrast to parametric methods, nonparametric density estimators do not explicitly restrict the space of considered PDFs. The most popular nonparametric method, kernel density estimation (KDE), places a symmetric density function $K(z)$, the so-called kernel, in each training data point $x_n$ \citep{Rosenblatt1956, Parzen1962}. 
The resulting density estimate for univariate distributions is formed as equally weighted mixture of the $N$ densities centered in the data points. In the case of multivariate kernel density estimation, i.e. $dim(\mathcal{X}) = l > 1$, the density can be estimated as product of marginal kernel density estimates. Such a kernel density estimate reads as follows:
\begin{equation}
    \hat{p}(\rx) = \prod_{j=1}^l \hat{p}(x^{(j)}) = \prod_{j=1}^l \frac{1}{N h^{(j)}} \sum_{n=1}^N K \left( \frac{x^{(j)} - x^{(j)}_n}{h^{(j)}} \right) \label{eq:kde}
\end{equation}
In that, $x^{(j)}$ denotes the j-th element of the column vector $\rx \in \mathcal{X} \subseteq \mathbb{R}^l$ and $h^{(j)}$ the bandwidth / somoothing parameter corresponding to the j-th dimension.

One popular choice of $K(\cdot)$ is the Gaussian kernel: $K(z) = (2\pi)^{-\frac{1}{2}} e^{-\frac{z^2}{2}}$
Other common choices of $K(\cdot)$ are the Epanechnikov and exponential kernels. Provided a continuous kernel function, the estimated PDF in (\ref{eq:kde}) is continuous. Beyond the appropriate choice of $K(\cdot)$, a central challenge is the selection of the bandwidth parameter $h$ which controls the smoothing of the estimated PDF. For details on bandwidth selection, we refer the interested reader to \citet{Li2007}.

%%%%%%%%%%%%%%%%%%%%%%%%%%%%%%%%%%%%%%%%%%%%%%%%%%%%%%%%%%%%%%%%%%%%%%%%%%%%%%%%%%%
\subsection{Conditional Density Estimation (CDE)} \label{background_cde}
Let $(X, Y)$ be a pair of random variables with respective domains $\mathcal{X} \subseteq \mathbb{R}^l$ and $\mathcal{Y} \subseteq \mathbb{R}^m$ and realizations $\rx$ and $\ry$. Let $p(\ry|\rx) = p(\rx,\ry) / p(\rx)$ denote the conditional probability density of $\ry$ given $\rx$. Typically, $Y$ is referred to as a dependent variable (i.e. explained variable) and $X$ as conditional (explanatory) variable. Given a dataset of observations $\mathcal{D} = \{(\rx_n, \ry_n)\}_{n=1}^N$ drawn from the joint distribution $(\rx_n, \ry_n) \sim p(\rx,\ry)$, the aim of conditional density estimation (CDE) is to find an estimate $\hat{p}(\ry|\rx)$ of the true conditional density $p(\ry|\rx)$.

In the context of conditional density estimation, the Kullback-Leibler divergence objective is expressed as expectation over $p(\rx)$:
\begin{equation}
    \mathbb{E}_{\rx \sim p(\rx)} \left[ \mathcal{D}_{KL}(p(\ry|\rx)||\hat{p}(\ry|\rx))\right] =   \mathbb{E}_{(\rx, \ry) \sim p(\rx, \ry)} \left[ \log p(\ry| \rx) - \log \hat{p}(\ry| \rx) \right] \label{eq:conditional_KL}
\end{equation} 
Similar to the unconditional case in (\ref{eq:mle-to-kl1}) - (\ref{eq:mle-to-kl2}),  parametric maximum likelihood estimation following from (\ref{eq:conditional_KL}) can be expressed as
\begin{equation}
    \theta^* = \text{arg} \max_\theta \sum_{n=1}^N \log \hat{p}_\theta(\ry_n | \rx_n) \label{eq:conditional_mle}
\end{equation}
Given a dataset $\mathcal{D}$ drawn i.i.d from $p(\rx, \ry)$, (\ref{eq:conditional_mle}) can be viewed as equivalent to minimizing a monte-carlo estimate of the expectation in (\ref{eq:conditional_KL}).
The nonparametric KDE approach, discussed in Section \ref{sec:nonparametric} can be extended to the conditional case. Typically, unconditional KDE is used to estimate both the joint density $\hat{p}(\rx, \ry)$ and the marginal density $\hat{p}(\rx)$. Then, the conditional density estimate follows as the density ratio
\begin{equation}
    \hat{p}(\ry| \rx) = \frac{\hat{p}(\rx, \ry)}{\hat{p}(\rx)}
\end{equation}
where both the enumerator and denominator are the sums of Kernel functions as in (\ref{eq:kde}). For more details on CDE, we refer the interested reader to \citet{Li2007}.

\section{Related Work}
This chapter discusses related work in the areas of finance, econometrics and machine learning. In that, we use the the differentiation between parametric and non-parametric methods, as discussed in Section \ref{sec:background}. In particular, we organize the following review in three categories: 1) parametric conditional density and time-series models with narrowly defined parametric families 2) non-parametric density estimation and 3) parametric models based on high-capacity function approximators such as neural networks.

\textbf{Parametric CDE in finance and econometrics.}
The majority of work in finance and econometrics uses a standard parametric family to model the conditional distribution of stock returns and other instruments. Typically, a Gaussian distribution is employed, whereby the parameters of the conditional distribution are predicted with time-series models. A popular instantiation of this category is the ARMA-GARCH method which models mean and variance of the conditional Gaussian through linear relationships \citep{Engle1982, Hamilton1994}. Various generalizations of GARCH attempt to model asymmetric return distributions and negative skewness \citep{Nelson1992, Glosten1993, Sentana1995}. Further work employs the student-t distribution as conditional probability model \citep{Bollerslev1987, Hansen1994} and models the dependence of higher-order moments on the past \citep{Gallant1991, Hansen1994}.

Though the neural network based CDE approaches, which are presented in this paper, are also parametric models, they make very little assumptions about the underlying relationships and density family. Both the relationship between the conditional variables and distribution parameters, as well as the probability density itself are modelled with flexible function classes (i.e. neural network and GMM). In contrast, traditional financial models impose strong assumptions such as linear relationships and Gaussian conditional distributions. It is unclear to which degree such modelling restrictions are consistent with the empirical data and how much they bias the inference.

\textbf{Non-parametric CDE.} 
A distinctly different line of work in econometrics aims to estimate densities in a non-parametric manner. Originally introduced in \citet{Rosenblatt1956, Parzen1962}, KDE uses kernel functions to estimate the probability density at a query point, based on the distance to all training points. In principle, kernel density estimators can approximate arbitrary probability distributions without parametric assumptions. However, in practice, data is finite and smoothing is required to achieve satisfactory generalization beyond the training data. The fundamental issue of KDE, commonly referred to as the bandwidth selection problem, is choosing the appropriate amount of smoothing \citep{Park1990, Cao1994}. Common bandwidth selection methods include rules-of-thumb \citep{Silverman1982, Sheather1991, Botev2010} and selectors based on cross-validation \citep{Rudemo1982, BOWMAN1984, Hall1992}.

In order to estimate conditional probabilities, previous work proposes to estimate both the joint and marginal probability separately with KDE and then computing the conditional probability as their ratio \citep{Hyndman1996, DeGooijer2003, Li2007}. Other approaches combine non-parametric elements with parametric elements \citep{Tresp2001, Sugiyama2010}, forming semi-parametric conditional density estimators. Despite their theoretical appeal, non-parametric density estimators suffer from the following drawbacks: First, they tend to generalize poorly in regions where data is sparse which especially becomes evident in the tail regions of the distribution. Second, their performance deteriorates quickly as the dimensionality of the dependent variable increases. This phenomenon is commonly referred to as the "curse of dimensionality". 

\textbf{CDE with neural networks: MDN, KDE, Normalizing Flows.}
The third line work approaches conditional density estimation from a parametric perspective. However, in contrast to parametric modelling in finance and econometrics, such methods use high-capacity function approximators instead of strongly constrained parametric families. Our work builds upon the work of \citet{Bishop1994} and \citet{Ambrogioni2017}, who propose to use a neural network to control the parameters of mixture density model. When both the neural network and the mixture of densities is chosen to be sufficiently expressive, any conditional probability distribution can be approximated \citep{Hornik1991, Li2000}. \citet{Sarajedini1999} propose neural networks that parameterize a generic exponential family distribution. However, this limits the overall expressiveness of the conditional density estimator. 

A recent trend in machine learning is the use of neural network based latent density models \citep{Mirza2014, Sohn2015}. Although such methods have been shown successful for estimating distributions of images, it is not possible to recover the PDF of such latent density models. More promising in this sense are normalizing flows which use a sequence of invertible maps to transform a simple latent distribution into more complex density functions \citep{Rezende2015a, Dinh2017, Trippe2018}. Since the PDF of normalizing flows is tractable, this could be an interesting direction to supplement our work.

While neural network based density estimators make very little assumptions about the underlying density, the suffer from severe over-fitting when trained with the maximum likelihood objective. In order to counteract over-fitting, various regularization methods have been explored in the literature \citep{Krogh1992, Holmstrom1992, Webb1994, Srivastava2014a}. However, these methods were developed with emphasis on regression and classification problems. Our work focuses on the regularization of neural network based density estimators. In that, we make use of the noise regularization framework \citep{Webb1994, Bishop1995}, discussing its implications in the context of density estimation and empirically evaluating its efficacy.

\section{Conditional Density Estimation with Neural Networks} \label{sec:Model}
The following chapter introduces and discusses two neural network based approaches for estimating conditional densities. Both density estimators are, in their nature, parametric models, but exhibit substantially higher flexibility than traditional parametric methods. In the first part, we formally define the density models and explain their fitting process. The second part of this chapter attends to the challenges that arise from this flexibility, introducing a form of smoothness regularization to combat over-fitting and enable good generalization.

%%%%%%%%%%%%%%%%%%%%%%%%%%%%%%%%%%%%%%%%%%%%%%%%%%%%%%%%%%%%%%%%%%%%%%%%%%%%%%%%%%%%%%%%%%%%%%%%
\subsection{The Density Models}

\subsubsection{Mixture Density Networks}
\label{mdn}
Mixture Density Networks (MDNs) combine conventional neural networks with a mixture density model for the purpose of estimating conditional distributions $p(\ry|\rx)$ \citep{Bishop1994}. In particular, the parameters of the unconditional mixture distribution $p(\ry)$ are outputted by the neural network, which takes the conditional variable $\rx$ as input. 

For our purpose, we employ a Gaussian Mixture Model (GMM) with diagonal covariance matrices as density model. The conditional density estimate $\hat{p}(\ry|\rx)$ follows as weighted sum of $K$ Gaussians
\begin{equation}
    \hat{p}(\ry|\rx) = \sum_{k=1}^K w_{k}(\rx;\theta) \mathcal{N}(\ry|\mu_{k}(\rx;\theta), \sigma_{k}^2(\rx;\theta))
\end{equation}
wherein $w_{k}(\rx;\theta)$ denote the weight, $\mu_{k}(\rx;\theta)$ the mean and $\sigma_{k}^2(\rx;\theta)$ the variance of the k-th Gaussian component. All the GMM parameters are governed by the neural network with parameters $\theta$ and input $\rx$. It is possible to use a GMM with full covariance matrices $\Sigma_k$ by having the neural network output the lower triangular entries of the respective Cholesky decompositions $\Sigma_k^{1/2}$ \citep{Tansey2016}. However, we choose diagonal covariance matrices in order to avoid the quadratic increase in the neural network's output layer size as the dimensionality of $\mathcal{Y}$ increases.

The mixing weights ${w_{k}(\rx;\theta)}$ must resemble a categorical distribution, i.e. it must hold that $\sum_{k=1}^K w_{k}(\rx;\theta)=1$ and $w_{k}(\rx;\theta)  \geq 0 ~ \forall k$. To satisfy the conditions, the softmax function is used.
\begin{equation}
   w_k(\rx) = \frac{\exp(a_k^{w}(\rx))}{\sum_{i=1}^K \exp(a_i^{w}(\rx))}
\end{equation}
In that, $a_k^{w}(\rx) \in \mathbb{R}$ denote the logit scores emitted by the neural network. Similarly, the standard deviations $\sigma_k(\rx)$ must be positive. To ensure that the respective neural network satisfy the non-negativity constraint, a sofplus non-linearity is applied: 

\begin{equation}
   \sigma_k(\rx) = \log \left( 1 + exp(a^{\sigma}_k(\rx)) \right)
\end{equation}

Since the component means $\mu_{k}(\rx;\theta)$ are not subject to such restrictions, we use a linear layer without non-linearity for the respective output neurons.

%%%%%%%%%%%%%%%%%%%%%%%%%%%%%%%%%%%%%%%%%%%%%%%%%%%%%%%%%%%%%%%%%%%%%%%%%%%%%%%%%%%%%%%%%%%%%%%
\subsubsection{Kernel Mixture Networks}
\label{kmn}
While MDNs resemble a purely parametric conditional density model, a closely related approach, the Kernel Mixture Network (KMN), combines both non-parametric and parametric elements \citep{Ambrogioni2017}. Similar to MDNs, a mixture density model of $\hat{p}(y)$ is combined with a neural network which takes the conditional variable $\rx$ as an input. However, the neural network only controls the weights of the mixture components while the component centers and scales are fixed w.r.t. to $\rx$.  Figuratively, one can imagine the neural network as choosing between a very large amount of pre-existing kernel functions to build up the final combined density function. As common for non-parametric density estimation, the components/kernels are placed in each of the training samples or a subset of the samples. For each of the kernel centers, one ore multiple scale/bandwidth parameters $\sigma_m$ are chosen. As for MDNs, we employ Gaussians as mixture components, wherein the scale parameter directly coincides with the standard deviation.

Let $K$ be the number of kernel centers $\mu_k$ and $M$ the number of different kernel scales $\sigma_m$. The KMN conditional density estimate reads as follows:

\begin{equation}
    \hat{p}(\ry|\rx) = \sum_{k=1}^K \sum_{m=1}^M w_{k,m}(\rx; \theta) \mathcal{N}(\ry|\mu_k, \sigma_m^2)
\end{equation}

As previously, the weights $w_{k,m}$ correspond to a softmax function. \citet{Ambrogioni2017} propose to choose the kernel centers $\mu_k$ by subsampling the training data by recursively removing each point $y_n$ that is closer than a constant $\delta$ to any of its predecessor points. This can be seen as a naive form of clustering which depends on the ordering of the dataset. Instead, we suggest to use a well-established clustering method such as K-means for selecting the kernel centers. The scales of the Gaussian kernels can either be fixed or jointly trained with the neural network weights. In practice, considering the scales $\{\sigma_m\}_{m=1}^M$ as trainable parameters consistently improves the performance.

Overall, the KMN model is more restrictive than MDN as the locations and scales of the mixture components are fixed during inference and cannot be controlled by the neural network. However, due to the reduced flexibility of KMNs, they are less prone to over-fit than MDNs.

\subsection{Fitting the Density Models} \label{sec:fitting}
The parameters $\theta$ of the neural network are fitted through maximum likelihood estimation. In practice, we minimize the negative conditional log-likelihood of the training data $\mathcal{D} = \{(\rx_n, \ry_n)\}_{n=1}^N$:
\begin{equation}
    \theta^* = \text{arg} \min_\theta - \sum_{n=1}^N \log p_\theta(\ry_n | \rx_n) \label{eq:max_likelihood}
\end{equation}

In that, the negative log-likelihood in (\ref{eq:max_likelihood}) is minimized via numerical optimization. Due to its superior performance in non-convex optimization problems, we employ stochastic gradient descent in conjunction with the adaptive learning rate method Adam \citep{Kingma2015}. 

%%%%%%%%%%%%%%%%%%%%%%%%%%%%%%%%%%%%%%%%%%%%%%%%%%%%%%%%%
\subsubsection{Variable Noise as Smoothness Regularization}
\label{sec:noise_reg}
A central issue when training high capacity function approximators such as neural networks is determining the optimal degree of complexity of the model. Models with too limited capacity may not be a able to sufficiently capture the structure of the data, inducing a strong restriction bias. On the other hand, if a model is too expressive it is prone to over-fit the training data, resulting in poor generalization. This problem can be regarded as finding the right balance when trading off variance against inductive bias.
There exist many techniques allowing to control the trade-off between bias and variance, involving various forms of regularization and data augmentation. For an overview overview of regularization techniques, the interested reader is referred to \cite{Kukacka2017}. 

% To a large degree, the contemporary practice of machine learning can be viewed as the art of carefully engineering the right inductive bias for the problem at hand. This means, using prior domain knowledge to select the right regularization terms and data augmentation methods, attempting to minimize the variance of the learning algorithm while not imposing biases that guide the learner away from good hypotheses.

Adding noise to the data during training can be viewed as form of data augmentation and regularization that biases towards smooth functions \citep{Webb1994, Bishop1994}. In the domain of finance, assuming smooth return distribution is a reasonable assumption. Hence, it is desirable to embed an inductive bias towards smoothness into the learning procedure in order to reduce the variance. 
Specifically, we add small perturbances in form of a random vector $\rxi \sim q(\rxi)$ to the data $\tilde{\rx}_n = \rx_n + \rxi_x$ and  $\tilde{\ry}_n = \ry_n + \rxi_y$. Further, we assume that the noise is zero centered as well as identically and independently distributed among the dimensions, with standard deviation $\eta$:
\begin{equation} \label{eq:nois_assumpt}
\E_{\rxi \sim  q(\rxi)}\left[ \rxi \right] = 0 ~~~ \text{and} ~~~ \E_{\rxi \sim  q(\rxi)}\left[  \rxi  \rxi^\top \right] = \eta^2 I
\end{equation}
Before discussing the particular effects of randomly perturbing the data during conditional maximum likelihood estimation, we first analyze noise regularization in a more general case. Let $\mathcal{L}_{\mathcal{D}}(\mathcal{D})$ be a loss function over a set of data points $\mathcal{D}=\{x_1, ..., x_N\}$, which can be partitioned into a sum of losses corresponding to each data point $x_n$:
\begin{equation}
    \mathcal{L}_{\mathcal{D}}(\mathcal{D}) = \sum_{n=1}^N \mathcal{L}(\rx_n)
\end{equation}
The loss $\mathcal{L}(\rx_n + \rxi)$, resulting from adding random perturbations can be approximated by a second order Taylor expansion around $x_n$ 
\begin{equation}
\mathcal{L}(\rx_n + \rxi) = \mathcal{L}(\rx_n) + \rxi^\top \nabla_{\rx} \mathcal{L}(\rx) \big \rvert_{\rx_n} + \frac{1}{2} \rxi^\top \nabla_\rx^2 \mathcal{L}(x) \big \rvert_{\rx_n} \rxi+ \mathcal{O}(\rxi^3) \label{eq_append:noise_reg_taylor}
\end{equation}
Assuming that the noise $\rxi$ is small in its magnitude, $\mathcal{O}(\rxi^3)$ is neglectable.
Using the assumption about $\rxi$ in (\ref{eq:nois_assumpt}), the expected loss an be written as
\begin{equation}
\E_{\rxi \sim  q(\rxi)}\left[\mathcal{L}(\rx_n + \rxi) \right] \approx \mathcal{L}(\rx_n) + \frac{1}{2} \E_{\rxi \sim  q(\rxi)}\left[ \rxi^\top \mathbf{H}^{(n)} \rxi  \right] =  \mathcal{L}(\rx_n) + \frac{\eta^2}{2} \text{tr}(\mathbf{H}^{(n)}) \label{eq:noise_reg_taylor}
\end{equation}
where $\mathcal{L}(\rx_n)$ is the loss without noise and $\mathbf{H}^{(n)}=\frac{\partial^2 \mathcal{L}}{\partial x^2}(x) \big \rvert_{x_n}$ the Hessian of $\mathcal{L}$ w.r.t $x$, evaluated at $x_n$. This result has been obtained earlier by \citet{Webb1994} and \citet{Bishop1994}. See Appendix\ref{appendix:noise_reg} for derivations. 

Previous work \citep{Webb1994, Bishop1994, An1996} has introduced noise regularization for regression and classification problems. However, to our best knowledge, noise regularization has not been used in the context of parametric density estimation. In the following, we derive and analyze the effect of noise regularization w.r.t. maximum likelihood estimation of conditional densities.

When concerned with maximum likelihood estimation of a conditional density $p_\theta (y|x)$, the loss function coincides with the negative conditional log-likelihood $\mathcal{L}(y_n, x_n) = - \log p(y_n|x_n)$. Let the standard deviation of the additive data noise $\rxi_x$, $\rxi_y$ be $\eta_x$ and $\eta_y$ respectively. Maximum likelihood estimation (MLE) with data noise is equivalent to minimizing the loss
\begin{align}
\mathcal{L}(\mathcal{D}) &\approx - \sum_{n=1}^N \log p_\theta(\ry_n|\rx_n) + \sum_{n=1}^N \frac{\eta^2_y}{2} \text{tr}(\mathbf{H}_\ry^{(n)}) + \sum_{n=1}^N \frac{\eta^2_\rx}{2} \text{tr}(\mathbf{H}_{\rx}^{(n)}) \label{eq:smoothness_reg}\\
&= - \sum_{n=1}^N \log p_\theta(\ry_n|\rx_n) - \frac{\eta^2_\ry}{2} \sum_{n=1}^N \sum_{j=1}^m \frac{\partial^2 \log p_\theta(\ry|\rx)}{\partial y^{(j)} \partial y^{(j)}} \big \vert_{\ry = \ry_n} -\frac{\eta^2_\rx}{2} \sum_{n=1}^N \sum_{j=1}^l \frac{\partial^2 \log p_\theta(\ry|\rx)}{\partial x^{(j)} \partial x^{(j)}} \big \vert_{\rx = \rx_n}
\end{align}
In that, the first term corresponds to the standard MLE objective while the other two terms constitute a smoothness regularization. The second term in (\ref{eq:smoothness_reg}) penalizes large negative second derivatives of the conditional log density estimate $\log p_\theta(\ry|\rx)$ w.r.t. $\ry$. As the MLE objective pushes the density estimate towards high densities and strong concavity in the data points $\ry_n$, the regularization term counteracts this tendency to over-fit and overall smoothes the fitted distribution. The third term penalizes large negative second derivatives w.r.t. the conditional variable $\rx$, thereby regularizing the sensitivity of the density estimate on changes in the conditional variable. This smoothes the functional dependency of $p_\theta(\ry|\rx)$ on $\rx$. As stated previously, the intensity of the smoothness regularization can be controlled through the standard deviation ($\eta_x$ and $\eta_y$) of the perturbations. 

\begin{figure}
    \centering
    \includegraphics[width=1.0\textwidth]{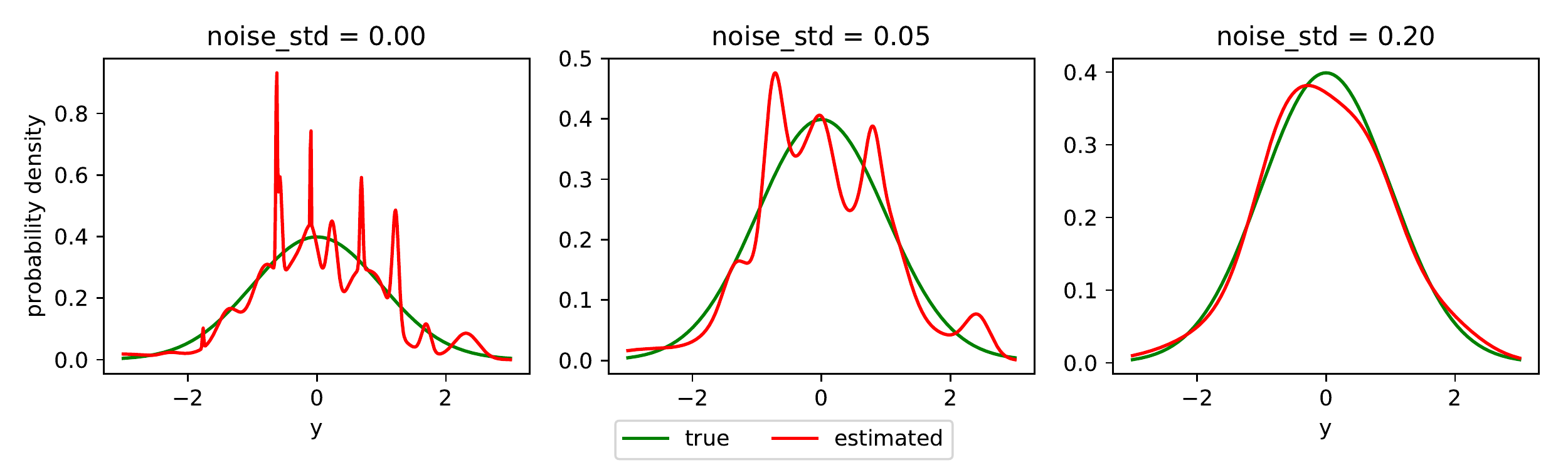}
    \caption{\textbf{Effect of noise regularization on density estimate.} Conditional MDN density estimate (red) and true conditional density (green) for different noise regularization intensities $\eta_\rx, \eta_\ry$. The MDN has been fitted with 3000 samples.}
    \label{fig:noise_reg_mdn}
\end{figure}

Figure \ref{fig:noise_reg_mdn} illustrates the effect of the introduced noise regularization scheme on MDN density estimates. Plain maximum likelihood estimation (left) leads to strong over-fitting, resulting in a spiky distribution that generalizes poorly beyond the training data. In contrast, training with noise regularization (center and right) results in smoother density estimates that are closer to the true conditional density. In Section \ref{sec:noise_reg_exps}, a comprehensive empirical evaluation demonstrates the efficacy and importance of noise regularization.

%%%%%%%%%%%%%%%%%%%%%%%%%%%%%%%%%%%%%%%%%%%%%%%%%%%%%%%%%%%%%%%%%%%%%%
%%%%%%%%%%%%%%%%%%%%%%%%%%%%%%%%%%%%%%%%%%%%%%%%%%%%%%%%%%%%%%%%%%%%%%
\subsubsection{Data Normalization} \label{sec:data_normalization}
In many applications of machine learning and econometrics, the value range of raw data varies widely. Significant differences in scale and range among features can lead to poor performance of many learning algorithms. When the initial distribution during training is statistically too far away from the actual data distribution, the training converges only slowly or might fail entirely. Moreover, the effect of many hyperparameters is often influenced by the value range of learning features and targets. For instance, the efficacy of noise regularization, introduced in the previous section, is susceptible to varying data ranges. In order to circumvent these and many more issues that arise due to different value ranges of the data, a common practice in machine learning is to normalize the data so that it exhibits zero mean and unit variance \citep{Sola1997, Grus2015}. While this practice is straightforward for classification and regression problems, such a transformation requires further consideration in the context of density estimation. The remainder of this section, elaborates on how to properly perform data normalization for estimating conditional densities. In that, we view the data normalization as change of variable and derive the respective density transformations that are necessary to recover an estimate of the original data distribution.

% Let $\mathcal{D}=\{(\rx_n,\ry_n)\}_{n=1}^N$ be a dataset where the tuples $(\rx_n, \ry_n) \sim p(\rx,\ry)$ are drawn from a joint distribution with density function $p: \mathbb{R}^{l} \times \mathbb{R}^{m} \rightarrow \mathbb{R}^+$. 
In order to normalize the training data $\mathcal{D}$ which originates from $p(\rx, \ry)$, we estimate mean $\hat{\mu}$ and standard deviation $\hat{\sigma}$ along each data dimension followed by subtracting the mean from the data points and dividing by the standard deviation.
\begin{equation}
\tilde{\rx} = \text{diag}(\hat{\sigma}_x)^{-1}(\rx-\hat{\mu}_\rx) ~~ \text{and} ~~ \tilde{\ry} = \text{diag}(\hat{\sigma}_y)^{-1}(\ry-\hat{\mu}_\ry) \label{eq:data_normalization}
\end{equation}
The normalization operations in (\ref{eq:data_normalization}) are linear transformations of the data. Subsequently, the conditional density model is fitted on the normalized data, resulting in the estimated PDF $\hat{q}_\theta(\tilde{\ry}| \tilde{\rx})$.  

However, when performing inference, one is interested in an unnormalized density estimate $\hat{p}_\theta(\ry| \rx)$, corresponding to the conditional data distribution $p(\ry|\rx)$. Thus, we have to transform the learned distribution $\hat{q}_\theta(\tilde{\ry}| \tilde{\rx})$ so that it coincides with $p(\ry | \rx)$. In that, both the transformations $\rx \rightarrow \tilde{\rx}$ and $\ry \rightarrow \tilde{\ry}$ must be accounted for. 

The former is straightforward: Since the neural network is trained to receive normalized inputs $\tilde{\rx}$, it is sufficient to transform the original inputs $\rx$ to $\tilde{\rx} = \text{diag}(\hat{\sigma}_\rx)^{-1}(\rx-\hat{\mu}_\rx)$ before feeding them into the network at inference time.
In order to account for the linear transformation of $\ry$, we have to use the change of variable formula since the volume of the probability density is not preserved if $\sigma_y \neq 1$. The change of variable formula can be stated as follows.

\begin{theorem} \label{theorem:change_of_variable}
Let $\tilde{Y}$ be a continuous random variable with probability density function $q(\tilde{y})$, and let $Y = v(\tilde{Y})$ be an invertible function of $\tilde{Y}$ with inverse $\tilde{Y} = v^{-1}(Y)$. The probability density function $p(y)$ of Y is:
\begin{equation}
    p(y) = q(v^{-1}(y)) * \bigg | \frac{d}{d y} \left( v^{-1}(y) \right) \bigg | \label{eq:change_of_variable}
\end{equation}

\end{theorem}
In that, $\big | \frac{d}{d y} \left( v(y) \right)\big |$ is the determinant of the Jacobian of $v$ which is vital for adjusting the volume of $q(v^{-1}(y))$, so that $\int p(y) dy = 1$.
In case of the proposed data normalization scheme, $v$ is a linear function
\begin{equation}
    v^{-1}(\ry) = \text{diag}(\hat{\sigma}_y)^{-1}(\ry-\hat{\mu}_\ry)
\end{equation}
and, together with (\ref{eq:change_of_variable}), $\hat{p}_\theta$ follows as
\begin{equation}
\hat{p}_\theta(\ry|\rx) = \big | \text{diag}(\hat{\sigma}_y)^{-1}|\hat{q}_\theta(\tilde{\ry} \big | ~ \tilde{\rx}) = \frac{1}{ \prod_{j=1}^l  \hat{\sigma}_\ry^{(j)} } ~ \hat{q}_\theta(\tilde{\ry}| \tilde{\rx}) \label{eq:reverse_transform}
\end{equation}
The above equation provides a simple method for recovering the unnormalized density estimate from the normalized mixture density $\hat{q}_\theta(\tilde{\ry}| \tilde{\rx})$. 

Alternatively, we can directly recover the conditional mixture parameters corresponding to $p_\theta(\ry|\rx)$. Let $(\tilde{w}_k, ~ \tilde{\mu}_k, ~ \text{diag}(\tilde{\sigma}_k))$ be the conditional parameters of the GMM corresponding to $q(\tilde{y}| \tilde{x})$. 
Based on the change of variable formula, Theorem \ref{theorem:gmm_lin_trans} provides a simple recipe for re-parameterizing the GMM so that it reflects the unnormalized conditional density.
As special case of Theorem \ref{theorem:gmm_lin_trans}, with $\Sigma=\text{diag}(\tilde{\sigma})$ and $B=\text{diag}(\hat{\sigma}_\ry)$, the transformed GMM corresponding to $\hat{p}_\theta(\ry|\rx)$ has the following parameters:
\begin{align}
& w_k = \tilde{w_k} \label{eq:gmm_transform_1}\\
& \mu_k = \hat{\mu}_y + \text{diag}(\hat{\sigma}_y) \tilde{\mu_k}  \\
& \sigma_k = \text{diag}(\hat{\sigma}_y) ~ \tilde{\sigma_k} \label{eq:gmm_transform_3} ~~ .
\end{align}

\begin{theorem} \label{theorem:gmm_lin_trans}
Let $\rx \in \mathbb{R}^n$ be a continuous random variable following a Gaussian Mixture Model (GMM), this is $\rx \sim p(\rx)$ with
\begin{equation}
 p(\rx) = \sum_{k=1}^{K} w_k ~ \mathcal{N}(\mu_k, \Sigma_k) ~~.
\end{equation}
Any linear transformation $\rz = a + B\rx$ of $\rx \sim p(\rx)$ with $a \in \mathbb{R}^n$ and B being an invertible $n \times n$ matrix follows a Gaussian Mixture Model with density function
\begin{equation}
 p(\rz) = \sum_{k=1}^{K} w_k ~ \mathcal{N}(a + B\mu_k, B \Sigma_k B^\top) ~~.
\end{equation}
Proof. See Appendix\ref{appendix:data_norm}
\end{theorem}

Overall, the training process with data normalization includes the following steps:
\begin{enumerate}
\item Estimate empirical unconditional mean $\hat{\mu}_\rx$ , $\hat{\mu}_\ry$ and standard deviation $\hat{\sigma}_\rx$, $\hat{\sigma}_\ry$ of training data
\item Normalize the training data: $\{(\rx_n, \ry_n) \} \rightarrow \{(\tilde{\rx}_n, \tilde{\ry}_n) \}$
$$\tilde{\rx}_n = \text{diag}(\hat{\sigma}_x)^{-1}(\rx_n-\hat{\mu}_\rx) ~, ~~ \tilde{\ry}_n = \text{diag}(\hat{\sigma}_y)^{-1}(\ry_n-\hat{\mu}_\ry) , ~~ n=1, ..., N$$
\item Fit the conditional density model $\hat{q}_\theta(\tilde{\ry}| \tilde{\rx})$ using the normalized data
\item Transform the estimated density back into the original data space to obtain $\hat{p}_\theta(\ry|\rx)$. This can be done by either
    \begin{enumerate}
        \item directly transforming the mixture density $\hat{q}_\theta$ with the change of variable formula in (\ref{eq:reverse_transform}) \textbf{or}
        \item transforming the mixture density parameters outputted by the neural network according to (\ref{eq:gmm_transform_1})-(\ref{eq:gmm_transform_3})
    \end{enumerate}
\end{enumerate}

\section{Empirical Evaluation with Simulated Densities}
\label{sec:sim_evaluation}
This chapter comprises an extensive experimental study based on simulated densities. It is organized as follows: In the first part, we explain the methodology of the experimental evaluation, including the employed conditional density simulations and evaluation metric. The following sections include an evaluation of the noise regularization and the data normalization scheme, which have been introduced in section \ref{sec:fitting}. Finally, we present a benchmark study, comparing the neural network based approaches with state-of-the art CDE methods.

\subsection{Methodology}
%%%%%%%%%%%%%%%%%%%%%%%%%%%%%%%%%%%%%%%%%%%%%%%%%%%%%%%%%%%%%%%%%%
\subsubsection{Density Simulation}
In order to benchmark the proposed conditional density estimators and run experiments that aim to answer different sets of questions, several data generating models (simulators) are employed. The density simulations allow us to generate unlimited amounts of data, and, more importantly, compute the statistical distance between the true conditional data distribution and the density estimate. 
The density simulations, briefly introduced below, are inspired by financial models and exhibit properties of empirical return distributions, such as negative skewness and excess kurtosis.

\begin{itemize}
    \item \textbf{EconDensity} (Fig. \ref{fig:econ_density}): Conditional Gaussian distribution that exhibits heteroscedasticity and a mean with quadratic dependence on the conditional variable.
    \item \textbf{ArmaJump} (Fig. \ref{fig:arma_jump}): AR(1) process with a jump component, thus exhibiting negative skewness and excess kurtosis.
    \item \textbf{SkewNormal} (Fig. \ref{fig:skew_norm}): Conditional skew normal distribution \citep{Andel1984}. The skewness, volatility and mean parameters of the conditional distribution are functions of the conditional variable. 
    \item \textbf{GaussianMixture} (Fig. \ref{fig:gmm}): The joint $p(\rx, \ry)$ distribution follows a GMM that can be factorized into a $x$- and $y$-component, i.e. $p(\rx, \ry)=p(\ry)p(\rx)$. This ensures that the conditional probability $p(\ry|\rx)$ is tractable. 
\end{itemize}
A detailed description of the conditional density simulations alongside an illustration (Figure \ref{fig:sim_densities}) of the respective probability density can be found in Appendix\ref{appendix:density_sim}.
%%%%%%%%%%%%%%%%%%%%%%%%%%%%%%%%%%%%%%%%%%%%%%%%%%%%%%%%%%%%%%%%%%%%%%%%%%%%%%%%%
\subsubsection{Evaluation Metrics}
In order to assess the goodness of the estimated conditional densities, we measure the statistical distance between the estimate and the true conditional probability density. In particular, the Hellinger distance is used as evaluation metric. We choose the Hellinger Distance over other popular statistical divergences, because it is symmetric and constrained to $[0,1]$, making it easier to interpret. Since the training data is simulated from a joint distribution $p(\rx,\ry)$, but the density estimates $\hat{p}(\ry|\rx)$ are conditional, we evaluate the statistical distance across different conditional values $\rx$. For that, we uniformly sample 10 values for $\rx$ between the 10\%- and 90\%-percentile of $p(\rx)$, compute the respective Hellinger distance and finally average the conditional statistical distances.
In all experiments, 5 different random seeds are used for the data simulation and density estimation. The reported Hellinger distances are averages over the random seeds while the  translucent areas in the plots depict the standard deviation across the seeds.

%%%%%%%%%%%%%%%%%%%%%%%%%%%%%%%%%%%%%%%%%%%%%%%%%%%%%%%%%%

\subsection{Evaluation of Noise Regularization} 
\label{sec:noise_reg_exps}

This section empirically examines the noise regularization, presented in Section \ref{sec:noise_reg}. As has been mathematically derived, the standard deviation of the variable noise controls the intensity of the smoothness regularization. Accordingly, the standard deviation parameters $\eta_x$ and $\eta_y$, can be used to trade-off between bias and variance of the conditional density estimator. 

\begin{figure}
    \centering
    \includegraphics[width=1.\textwidth]{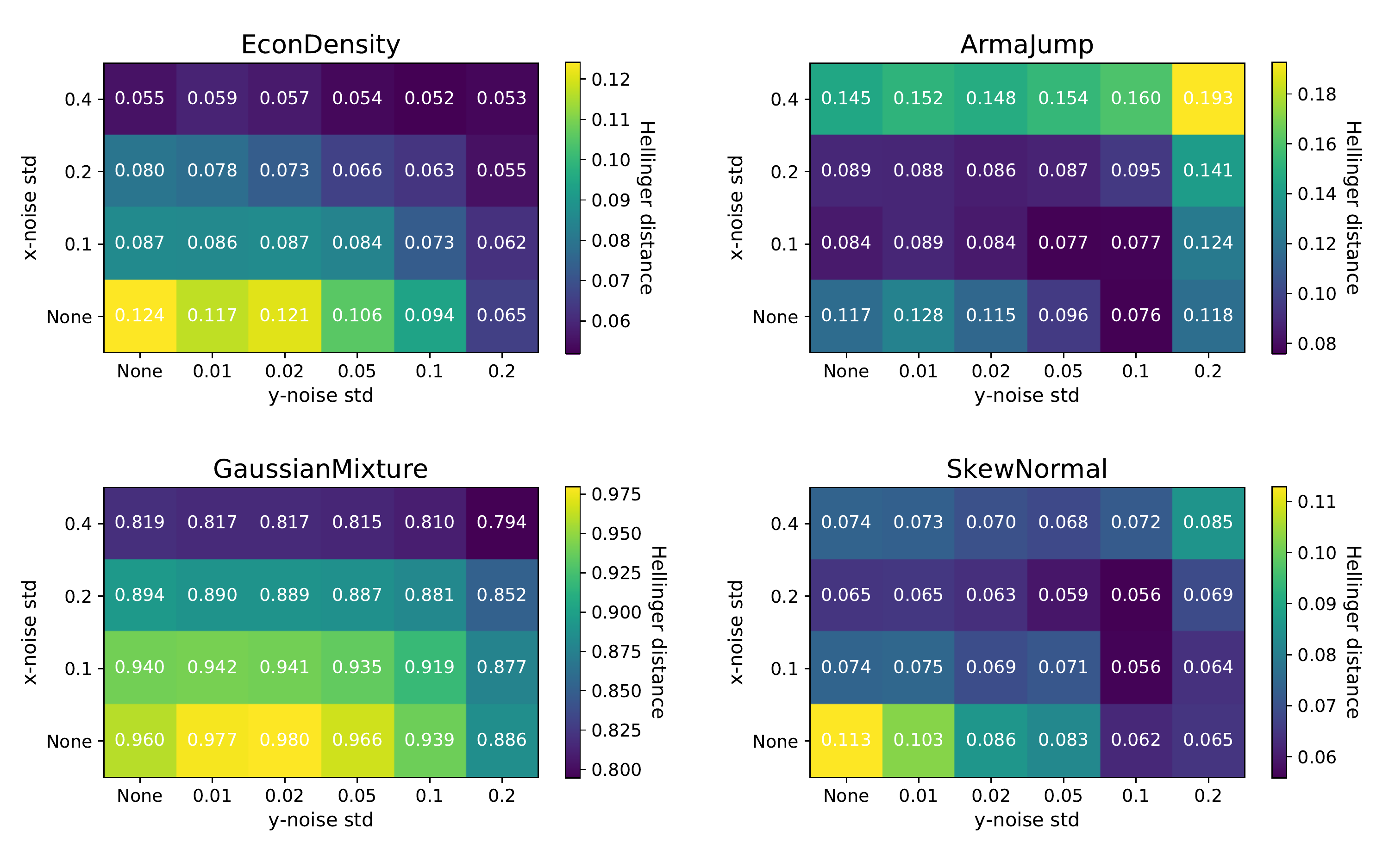}
    \caption{\textbf{Effect of various noise regularization intensities $\eta_x$ and $\eta_y$} Hellinger distance between MDN estimate, fitted with 1600 samples, and the true density across various noise regularization intensities. The displayed values are averages over 5 seeds.}
    \label{fig:xy_noise_headplot}
\end{figure}

In the following, we aim to empirically determine which values of $\eta_x$ and $\eta_y$ lead to the best density fit. For that, a grid-search on the noise regularization intensities is performed. Figure \ref{fig:xy_noise_headplot} holds the results in from of a heatplot. We observe that no regularization or noise regularization with low intensity leads to poor generalization of the density estimates, resulting in a high statistical distance w.r.t. the true density. In contrast, strong regularization may results in too much inductive bias, as it is the case for the ArmaJump simulation. The configuration $\eta_x = 0.2, \eta_y=0.1$ leads to low statistical distances in all density simulations. Hence, we choose it as default configuration for further experiments.

\begin{figure}
    \centering
    \includegraphics[width=1.\textwidth]{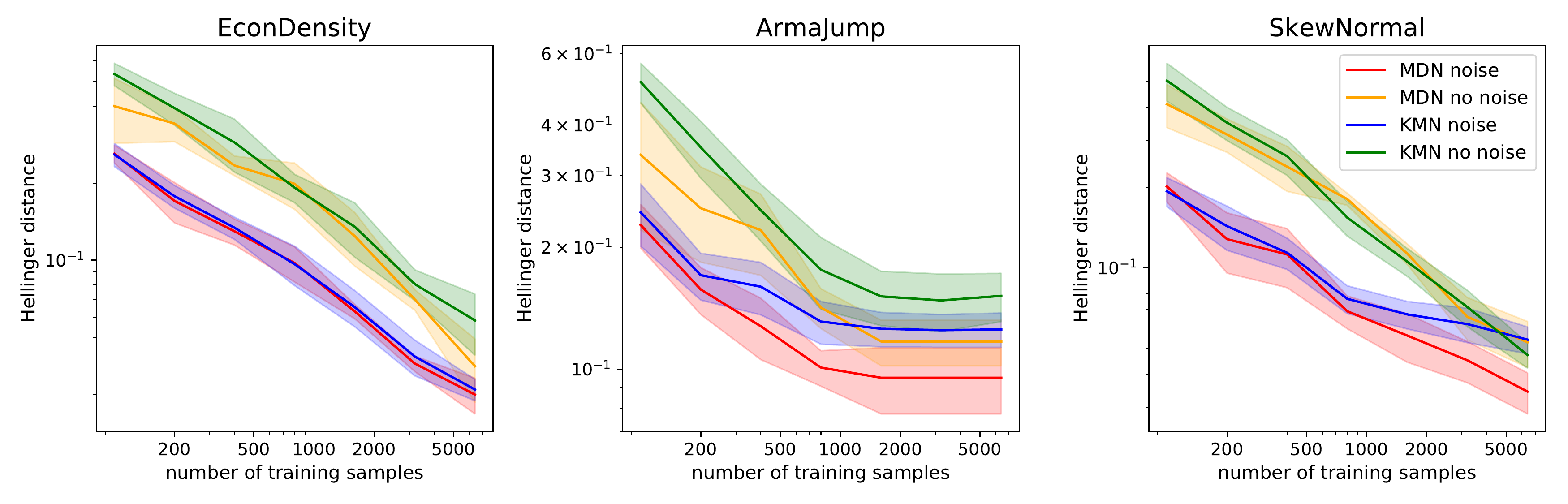}
    \caption{\textbf{Effect of noise regularization on goodness of estimated density.} Goodness of MDN/KMN density estimate, fitted with ($\eta_x=0.2, \eta_y=0.1$) and without noise regularization. The colored graphs display the Hellinger distance between estimated and true density, averaged over 5 seeds, and the translucent areas the respective standard deviation across varying samples sizes.}
    \label{fig:xy_noise_lineplot}
\end{figure}

Figure \ref{fig:xy_noise_lineplot} compares the goodness of KMN/MDN density estimates with and without noise regularization across different sizes of the training set. Consistent with Figure \ref{fig:xy_noise_headplot}, the noise regularization reduces the Hellinger distance by a significant margin. As the number of training samples increases, the estimator is less prone to over-fit and the positive effect of noise regularization tends to become smaller. Overall, the experimental results underpin the efficacy of noise regularization and its importance for achieving good generalization beyond the training samples.

%%%%%%%%%%%%%%%%%%%%%%%%%%%%%%%%%%%%%%%%%%%%%%%%%%%%%%%%%%%%%%%%%%%%%%

\subsection{Evaluation of Data Normalization} \label{sec:exp_data_norm}
\begin{figure}
    \centering
    \includegraphics[width=1.0\textwidth]{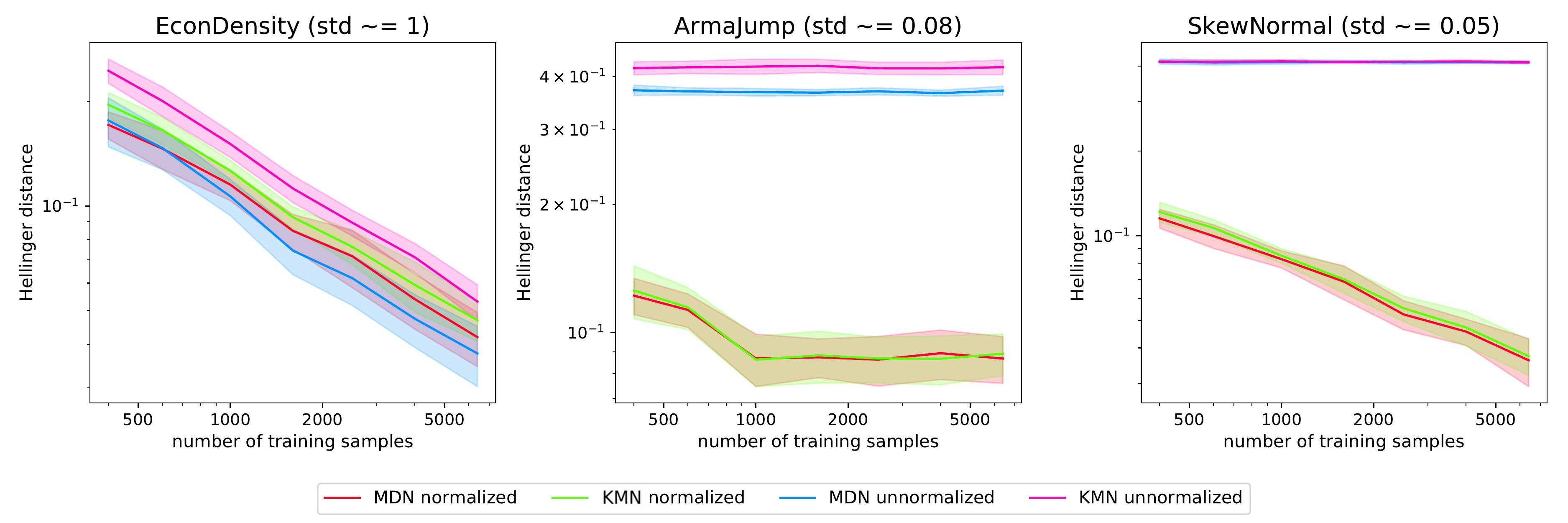}
    \caption{\textbf{Effect of data normalization.} Goodness of MDN / KMN density estimate, fitted with and without data normalization. The colored graphs display the Hellinger distance between estimated and true density, averaged over 5 seeds, and the translucent areas the respective standard deviation across varying samples sizes. While the EconDensity has a unconditional standard deviation of 1, the other two density simulations have a substantially lower unconditional volatility, ca. 0.08 and 0.05.}
    \label{fig:data_normalization}
\end{figure}

The data normalization scheme, introduced in section \ref{sec:data_normalization}, aims to make the hyperparameters of the density estimator invariant to the distribution of the training data. If no data normalization is used, the goodness of the hyperparameters is sensitive to the level and volatility of the training data. The most sensitive hyperparameter is the initialization of the parametric density estimator. If the initial density estimate is statistically too far away from the true density, numerical optimization might be slow or fail entirely to find a good fit. Figure \ref{fig:data_normalization} illustrates this phenomenon and emphasizes the practical importance of proper data normalization. 

In case of the EconDensity simulation, the conditional standard deviation of the simulation density and the initial density estimate are similar. Both density estimation with and without data normalization yield quite similar results. Yet, the data normalization consistently reduces the Hellinger distance. The ArmaJump and SkewNormal density simulators have substantially smaller conditional standard deviations, i.e. 12 - 20 times smaller than the EconDensity. Without the data normalization scheme, the initial KMN/MDN density estimates exhibit a large statistical distance to the true conditional density. As a result, the numerical optimization is not able to sufficiently fit the density within 1000 training epochs. As can be seen in Figure \ref{fig:data_normalization}, the resulting density estimates are substantially offset compared to the estimates with data normalization.

%%%%%%%%%%%%%%%%%%%%%%%%%%%%%%%%%%%%%%%%%%%%%%%%%%%%%%%%%%%%%%%%%%%%%

\subsection{Conditional Density Estimator Benchmark Study}
\label{sec:sim_benchmark}
After empirically evaluating the noise regularization and data normalization scheme, we benchmark the neural network based density estimators against state-of-the art conditional density estimation approaches. Specifically, the benchmark study comprises the following conditional density estimators:

\begin{itemize}
\item \textbf{Mixture Density Network (MDN):} As introduced in Section \ref{mdn}. The MDN is trained with data normalization and noise regularization ($\eta_x=0.2, \eta_y=0.1$).

\item \textbf{Kernel Mixture Network (KMN):} As introduced in Section \ref{kmn}. The KMN is trained with data normalization and noise regularization ($\eta_x=0.2, \eta_y=0.1$).

\item \textbf{Conditional Kernel Density Estimation (CKDE):} Non-parametric approach introduced in \ref{background_cde} using bandwidth selection via the rule-of-thumb \citep{Silverman1982}.

\item \textbf{CKDE with bandwidth selection via cross-validation (CKDE-CV):} Similar to CDKE using bandwidth selection via maximum likelihood cross-validation \citep{Li2007}

\item \textbf{$\epsilon$-Neighborhood kernel density estimation (NKDE):} Non-parametric method that considers only a local subset of training points to form a density estimate.
    
\item \textbf{Least-Squares Conditional Density Estimation (LSCDE):} Semi-parametric estimator that computes the conditional density as linear combination of kernels \citep{Sugiyama2010}.
\end{itemize}

\begin{figure}
    \centering
    \includegraphics[width=1.0\textwidth]{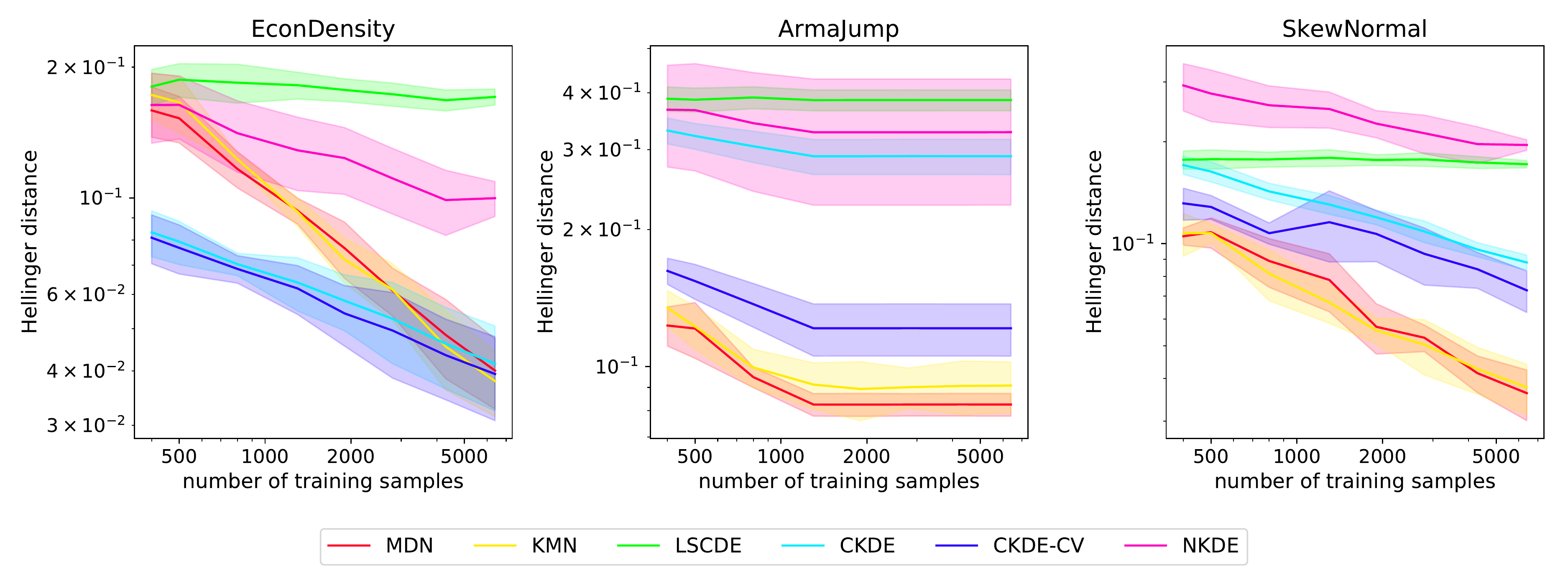}
    \caption{\textbf{Conditional Density Estimator Benchmark.} The illustrated benchmark study compares 6 density estimators across in 3 density simulations. To asses the goodness of fit, we report the Hellinger distance between the true density and the density estimate fitted with different sample sizes. The colored graphs display the Hellinger distance averaged over 5 seeds and the translucent areas the respective standard deviation.}
    \label{fig:sim_benchmark}
\end{figure}

Figure \ref{fig:sim_benchmark} depicts the evaluation results for the described estimators across different density simulations and number of training samples. Due to its limited modelling capacity, LSCDE yields poor estimates in all three evaluation cases and shows only minor improvements as the number of samples increases. CKDE consistently outperforms NKDE. This may be ascribed to the locality of the considered data neighborhoods of the training points that NKDE exhibits, whereas CKDE is able to fully use the available data. Unsurprisingly, the version of CKDE with bandwidth selection through cross-validation always improves upon CKDE with the rule-of-thumb.

In the EconDensity evaluation, CKDE achieves lower statistical distances for small sample sizes. However, the neural network based estimators KMN and MDN gain upon CKDE as the sample size increases and achieve similar results in case of 6000 samples. In the other two evaluation cases, KMNs and MDNs consistently outperform the other estimators. This demonstrates that even for small sample sizes, noise-regularized neural network estimators can be an equipollent or even superior alternative to well established CKDE.

%%%%%%%%%%%%%%%%%%%%%%%%%%%%%%%%%%%%%%%%%%%%%%%%%%%%%%%%%%%%%%%%%%%%%%%%%%%%%%%%
\section{Empirical Evaluation on Euro Stoxx 50 Data} \label{sec:eurostoxx_eval}
While the previous chapter is based on simulations, this chapter provides an empirical evaluation and benchmark study on real-world stock market data. In particular, we are concerned with estimating the conditional probability density of Euro Stoxx 50 returns. After a describing the data and density estimation task in detail, we report and discuss benchmark results. 

%%%%%%%%%%%%%%%%%%%%%%%%%%%%%%%%%%%%%%%%%%%%%%%%%%%%%%%%%%%%%%%%%%%%%%%%%%%%%%%%
\subsection{The Euro Stoxx 50 data} \label{sec:eurostoxx_data}
The following empirical evaluation is based the Euro Stoxx 50 stock market index. The data comprises 3169 trading days, dated from January 2003 until June 2015. We define the task as predicting the conditional probability density of 1-day log returns, conditioned on 14 explanatory variables. These conditional variables include last period returns, risk free rate, realized volatility, option-implied moments and return factors. For a detailed description of the conditional variables, we refer to Appendix\ref{appendix:eurostoxx_data}.
Overall, the target variable is one-dimensional, i.e. $\ry \in \mathcal{Y} \subseteq \mathbb{R}$, whereas the conditional variable $\rx$ constitutes a 14-dimensional vector, i.e. $\rx \in \mathcal{X} \subseteq \mathbb{R}^{14}$.

%%%%%%%%%%%%%%%%%%%%%%%%%%%%%%%%%%%%%%%%%%%%%%%%%%%%%%%%%%%%%%%%%%%%%%%%%%%%%%%
\subsection{Evaluation Methodology}
In order to assess the goodness of the different density estimators, out-of-sample validation is used. In particular, the available data is split in a training set which has a proportion of 80 \% and a validation set consisting of the remaining 20 \% of data points. The validation set $\mathcal{D}_{val}$ is a consecutive series of data, corresponding to the 633 most recent trading-day and is only used for computing the following goodness-of-fit measures:

\begin{itemize}
\item \emph{Avg. log-likelihood:} Average conditional log likelihood of validation data
\begin{equation}
\frac{1}{|\mathcal{D}_{val}|}\sum_{(\rx, \ry) \in \mathcal{D}_{val}} \log \hat{p}(\ry| \rx)
\end{equation}

\item \emph{RMSE mean:} Root-Mean-Square-Error (RMSE) between the realized log-return and the mean of the estimated conditional distribution. The estimated conditional mean is defined as the expectation of $\ry$ under the distribution $\hat{p}(\ry|\rx)$:
\begin{equation}
    \hat{\mu}(\rx) = \int_\mathcal{Y} \ry \, \hat{p}(\ry|\rx) d\ry
\end{equation}
Based on that, the RMSE w.r.t. $\hat{\mu}$  is calculated as
\begin{equation}
    RMSE_\mu = \sqrt{\frac{1}{|\mathcal{D}_{val}|}\sum_{(\rx, \ry) \in \mathcal{D}_{val}} \left( \ry - \hat{\mu}(\rx)\right)^2}
\end{equation}

\item \emph{RMSE Std:} RMSE between the realized deviation from the predicted mean $\hat{\mu}(\rx)$ and the standard deviation of the conditional density estimate. The estimated conditional standard deviation is defined as
\begin{equation}
    \hat{\sigma}(\rx) = \sqrt{\int_\mathcal{Y} (\ry - \hat{\mu}(\rx))^2 \hat{p}(\ry|\rx) d\ry}
\end{equation}
The respective RSME is calculated as follows:
\begin{equation}
    RMSE_\sigma = \sqrt{\frac{1}{|\mathcal{D}_{val}|}\sum_{(\rx, \ry) \in \mathcal{D}_{val}} \left( |\ry - \hat{\mu}(\rx)|  - \hat{\sigma}(\rx)\right)^2}
\end{equation}
\end{itemize}
For details on the estimated conditional moments and the approximation of the associated integrals, we refer the interested reader to Appendix\ref{appendix:compute_moments}.

Calculating the average log-likelihood is a common way of evaluating the goodness of a density estimate \citep{Rezende2015a, Tansey2016, Trippe2018}. The better the estimated conditional density approximates the true distribution, the higher the out-of-sample likelihood on expectation. Only if the estimator generalizes well beyond the training data, it can assign high conditional probabilities to the left-out validation data.

In finance, return distributions are often characterized by their centered moments. The RMSEs w.r.t. mean and standard deviation provide a quantitative measure for the predictive accuracy and consistency w.r.t. the predictive uncertainty.
Overall, the training of the estimators and calculation of the goodness measures is performed with 5 different seeds. The reported results are averages over the 5 seeds, alongside the respective standard deviation.

% We note that predicting the target variable might be subject to the following sources of error:
% \begin{itemize}
%     \item the neural network-based methods can introduce jumps when estimating returns, yielding that the Wiener process assumption cannot be satisfied in these cases which results in an estimator incorrectly estimating the variance
%     \item estimating the standard deviation of daily return might be incorrect because the Wiener process assumption does not hold
%     \item the intraday standard deviation estimation reflects the standard deviation of the unconditional return distribution. In contrast, we estimate conditional densities, yielding on average a smaller standard deviation than the unconditional density
% \end{itemize}
%%%%%%%%%%%%%%%%%%%%%%%%%%%%%%%%%%%%%%%%%%%%%%%%%%%%%%%%%%%%%%%%%%%%%%%
\subsection{Empirical Density Estimator Benchmark}

\begin{table}
\centering
\begin{tabular}{l|c|c|c}
{} &  Avg. log-likelihood &    RMSE mean ($10^{-2}$)  &  RMSE std ($10^{-2}$) \\ \hline
CKDE &     3.3368 $\pm$ 0.0000 &   0.6924 $\pm$ 0.0000 & 0.8086 $\pm$ 0.0000   \\ 
NKDE &  3.1171 $\pm$ 0.0000 & 1.0681 $\pm$ 0.0000 & 0.5570 $\pm$ 0.0000  \\ 
LSCDE  &  3.5079  $\pm$  0.0087 &   0.7057 $\pm$ 0.0061 &  0.5442  $\pm$ 0.0028 \\ \hline
MDN w/o noise & 3.1386 $\pm$ 0.1501 &   \bf{0.5339 $\pm$ 0.0084} &  \bf{0.3222 $\pm$ 0.0064} \\
KMN w/o noise & 3.3130 $\pm$ 0.0743 &   0.6070 $\pm$ 0.0417 &  0.4072 $\pm$  0.0372 \\
MDN w/ noise  & \bf{3.7539 $\pm$  0.0324} &   \bf{0.5273 $\pm$ 0.0082} &  \bf{0.3188 $\pm$ 0.0016}  \\
KMN w/ noise  & \bf{3.7969 $\pm$  0.0250} &   0.5375 $\pm$ 0.0079 &  0.3254 $\pm$ 0.0063 \\
% LSCDE & 3.5072 $\pm$ 0.0021 & 0.7105 $\pm$ 0.0047 & 0.5451 $\pm$ 0.0029 \\ \hline
% MDN w/o noise   &  3.2797 $\pm$ 0.2058 & \bf{0.5279 $\pm$ 0.0075} & \bf{0.3185 $\pm$ 0.0048}  \\
% KMN w/o noise   &  3.3578 $\pm$ 0.0653 & 0.5903 $\pm$ 0.0339 & 0.3673 $\pm$ 0.0107  \\
% MDN w/ noise  &  \bf{3.7991 $\pm$ 0.0142} & \bf{0.5224 $\pm$ 0.0019} & \bf{0.3171 $\pm$ 0.0034}  \\
% KMN w/ noise &   \bf{3.8010 $\pm$ 0.0142} & 0.5342 $\pm$ 0.0062 & 0.3287 $\pm$ 0.0034 \\
\end{tabular}
\caption{$~$ \textbf{Out-of-sample validation on EuroStoxx 1-day returns.}}
\label{table:eurostoxx_eval}
\end{table}

\begin{table}
\centering
\begin{tabular}{l|c|c|c}
    {} &  Avg. log-likelihood &    RMSE mean ($10^{-2}$)  &  RMSE std ($10^{-2}$) \\ \hline
    CKDE LOO-CV &        3.8142 $\pm$ 0.0000 &  0.5344 $\pm$ 0.0000 &  0.3672 $\pm$ 0.0000\\
    NKDE LOO-CV &        3.3435 $\pm$ 0.0000 &  0.7943 $\pm$ 0.0000 &  0.4831 $\pm$ 0.0000 \\
    %LSCDE 10-fold CV & 3.5292 $\pm$ 0.0069 & 0.6803  $\pm$ 0.0049 & 0.5477  $\pm$ 0.0038 \\ \hline
    LSCDE 10-fold CV & 3.5250 $\pm$ 0.0040 & 0.6836  $\pm$ 0.0056 & 0.5538  $\pm$ 0.0059 \\ \hline
    MDN 10-fold CV &     \textbf{3.8351 $\pm$  0.0114} &    \textbf{0.5271 $\pm$       0.0060} & \textbf{ 0.3269 $\pm$ 0.0026}\\
    KMN 10-fold CV &        \textbf{3.8299 $\pm$ 0.0141} &   \textbf{0.5315 $\pm$ 0.0075} &  \textbf{0.3247 $\pm$ 0.0047}
    % MDN 10-fold CV &     \textbf{3.8354 $\pm$  0.0095} &    \textbf{0.5250 $\pm$       0.0075} & \textbf{ 0.3266 $\pm$ 0.0009}\\
    % KMN 10-fold CV &        \textbf{3.8270 $\pm$ 0.0162} &   \textbf{0.5327 $\pm$ 0.0080} &  \textbf{0.3308 $\pm$ 0.0062}
\end{tabular}
\caption{$~$ \textbf{Out-of-sample validation on EuroStoxx 1-day returns - Hyper-Parameter selection via cross-validation.}}
\label{table:eurostoxx_eval_cv}
\end{table}

The Euro Stoxx 50 estimator benchmark consists of two categories. 
First, we compare the estimators in their default hyper-parameter configuration. The default configurations have been selected with hyperparameter sweeps on the simulated densities in the previous chapters. The respective results are reported in Table \ref{table:eurostoxx_eval}. Consistent with the simulation studies in Section \ref{sec:sim_evaluation}, noise regularization for MDNs and KMNs improves the estimate's generalization (i.e. validation log-likelihood) by a significant margin. Moreover, the regularization improves the predictive accuracy (RMSE mean) and uncertainty estimates (RMSE std).

In the second part of the benchmark, the estimator parameters are selected through cross-validation on the Euro Stoxx training set. As goodness criterion, the log-likelihood is used. Following previous work \citep{Duin1976, Pfeiffer1985, Li2007}, we use leave-one-out cross-validation in conjunction with the downhill simplex method of \citet{Nelder1965} for selecting the parameters of the kernel density estimators (CKDE and NKDE). For the remaining methods, hyper-parameter grid search with 10-fold cross-validation is employed. For details on the determined hyper-parameter settings, we refer to Appendix\ref{appendix:hyperparam}. Table \ref{table:eurostoxx_eval_cv} depicts the evaluation results with hyper-parameter search. Similar to the results without hyper-parameter selection, MDNs and KMNs with noise regularization consistently outperform previous methods in all three evaluation measures. This strengthens the results from the simulation study and demonstrates that, when regularized properly, neural network based methods are able to generate superior conditional density estimates.

Moreover, it is interesting to observe that both kernel density estimators, improve substantially, when cross-validation is used. This is a strong indication, that the bandwidth which is selected through the Gaussian rule-of-thumb is inferior and the underlying return data is non-Gaussian.

\section{Conclusion}
This paper studies the use of neural networks for conditional density estimation. Addressing the problem of over-fitting, we introduce a noise regularization method that leads to smooth density estimates and improved generalization. Moreover, a normalization scheme which makes the model's hyper-parameters insensitive to differing value ranges is proposed. Corresponding experiments showcase the effectiveness and practical importance of the presented approaches. In a benchmark study, we demonstrate that our training methodology endows neural network based CDE with a better out-of-sample performance than previous semi- and non-parametric methods. Overall, this work establishes a practical framework for the successful application of neural network based CDE in areas such as econometrics. Based on the promising results, we are convinced that the proposed method enhances the econometric toolkit and thus advocate further research in this direction. While this paper focuses on CDE with mixture densities, a promising avenue for future research could be the use of normalizing flows as parametric density representation.

%%%%%%%%%%%%%%%%%%%% Appendix %%%%%%%%%%%%%%%%%%%%%%%%%%%%%%%%%
\clearpage
\appendix
\titleformat*{\subsubsection}{\normalfont}
\section*{Appendix}

%%%%%%%%%%%%%%%%%%%%%%%%%%%%%%%%%%%%%%%%%%%%%%%%%%%%%%%%%%%%%%%%%%%%%%%%%%%%%%%%%%%%%%%%%%%
\subsection{Noise Regularization} \label{appendix:noise_reg}
Let $\mathcal{L}_{\mathcal{D}}(\mathcal{D})$ be a loss function over a set of data points $\mathcal{D}=\{\rx_1, ..., \rx_N\}$, which can be partitioned into a sum of losses corresponding to each data point $x_n$:
\begin{equation}
    \mathcal{L}_{\mathcal{D}}(\mathcal{D}) = \sum_{i=1}^N \mathcal{L}(\rx_n)
\end{equation}
Also, let each $x_n$ be perturbed by a random noise vector $\rxi \sim q(\rxi)$ with zero mean and i.i.d. elements, i.e.
\begin{equation} \label{eq_append:nois_assumpt}
\E_{\rxi \sim  q(\rxi)}\left[ \rxi \right] = 0 ~~~ \text{and} ~~~ \E_{\rxi \sim  q(\rxi)}\left[  \rxi_n \rxi_j^\top \right] = \eta^2 I
\end{equation}
The resulting loss $\mathcal{L}(\rx_n + \rxi)$ can be approximated by a second order Taylor expansion around $x_n$ 
\begin{equation}
\mathcal{L}(\rx_n + \rxi) = \mathcal{L}(\rx_n) + \rxi^\top \nabla_\rx \mathcal{L}(\rx) \big \rvert_{\rx_n} + \frac{1}{2} \rxi^\top \nabla_x^2 \mathcal{L}(\rx) \big \rvert_{\rx_n} \rxi+ \mathcal{O}(\rxi^3) \label{eq_append:noise_reg_taylor}
\end{equation}
Assuming that the noise $\rxi$ is small in its magnitude $\mathcal{O}(\rxi^3)$ may be neglected. The expected loss under $q(\rxi)$ follows directly from (\ref{eq_append:noise_reg_taylor}):
\begin{equation}
\E_{\rxi \sim  q(\rxi)}\left[\mathcal{L}(\rx_n + \rxi) \right] = \mathcal{L}(\rx_n) + \E_{\rxi \sim  q(\rxi)}\left[ \rxi^\top \nabla_x \mathcal{L}(\rx) \big \rvert_{\rx_n} \right]   + \frac{1}{2} \E_{\rxi \sim  q(\rxi)}\left[ \rxi^\top \nabla_x^2 \mathcal{L}(\rx) \big \rvert_{\rx_n} \rxi  \right]   
\label{eq_append:noise_reg_taylor_expect}
\end{equation}
Using the assumption about $\rxi$ in (\ref{eq_append:nois_assumpt}) we can simplify (\ref{eq_append:noise_reg_taylor_expect}) as follows:
\begin{align}
   \E_{\rxi \sim  q(\rxi)}\left[\mathcal{L}(\rx_n + \rxi) \right] &= \mathcal{L}(\rx_n) + \E_{\rxi \sim  q(\rxi)}\left[ \rxi \right]^\top \nabla_x \mathcal{L}(\rx) \big \vert_{\rx_n}  + \frac{1}{2} \E_{\rxi \sim  q(\rxi)}\left[ \rxi^\top \nabla_x^2 \mathcal{L}(\rx) \big \rvert_{\rx_n} \rxi  \right]  \\
    &= \mathcal{L}(\rx_n) + \frac{1}{2} \E_{\rxi \sim  q(\rxi)}\left[ \rxi^\top \mathbf{H}^{(n)} \rxi  \right] \\
    &= \mathcal{L}(\rx_n) + \frac{1}{2} E_{\rxi \sim  q(\rxi)}\left[ \sum_j \sum_k \xi_j \xi_k \frac{\partial^2 \mathcal{L}(\rx)}{\partial x^{(j)} \partial x^{(k)}} \bigg \vert_{\rx_n}\right] \\
    &= \mathcal{L}(\rx_n) + \frac{1}{2}  \sum_j E_{\rxi}\left[ \xi_j^2 \right] \frac{\partial^2 \mathcal{L}(\rx)}{\partial x^{(j)} \partial x^{(j)}} \bigg \vert_{\rx_n} + \frac{1}{2} \sum_j \sum_{k \neq j} E_{\rxi}\left[ \xi_j \xi_k \right]  \frac{\partial^2 \mathcal{L}(\rx)}{\partial x^{(j)} \partial x^{(k)}} \bigg \vert_{\rx_n} \\
    &= \mathcal{L}(\rx_n) + \frac{\eta^2}{2}  \sum_j \frac{\partial^2 \mathcal{L}(\rx)}{\partial x^{(j)} \partial x^{(j)}} \bigg \vert_{\rx_n} \\
     &= \mathcal{L}(\rx_n) + \frac{\eta^2}{2}\text{tr}(\mathbf{H}^{(n)}) 
\end{align}
In that, $\mathcal{L}(\rx_n)$ is the loss without noise and $\mathbf{H}^{(n)} = \nabla_\rx^2 \mathcal{L}(\rx) \big \rvert_{\rx_n}$ the Hessian of $\mathcal{L}$ at $\rx_n$. With $\xi_j$ we denote the elements of the column vector $\rxi$.\\

%%%%%%%%%%%%%%%%%%%%%%%%%%%%%%%%%%%%%%%%%%%%%%%%%%%%%%%%%%%%%%%%%%%%%%%%%%%%%

\subsection{Data Normalization and Change of Variable} \label{appendix:data_norm}

\begin{lemma} \label{lemma:pdf_lin_trans_appendix}
Let $\rx \in S \subseteq \mathbb{R}^n$ be a continuous random variable with probability density function $p(\rx)$. Any linear transformation $\rz = \ra + B\rx$ of $\rx \sim p(\rx)$ with $\ra \in \mathbb{R}^n$ and B being an invertible $n \times n$ matrix follows the probability density function
\begin{equation}
q(\rz) = \frac{1}{|B|} ~ p\left( B^{-1}(\rx-\ra) \right) ~,~~ \rz \in \{a+B\rx ~ \rx\in S \} ~~.
\end{equation}
\end{lemma}

\begin{refproof}[Proof of Lemma \ref{lemma:pdf_lin_trans_appendix}]
The Lemma directly follows from the change of variable theorem (see \citet{Bishop2006} page 18)
\end{refproof}

\begin{theorem} \label{theorem:gmm_lin_trans_appendix}
Let $\rx \in \mathbb{R}^n$ be a continuous random variable following a Gaussian Mixture Model (GMM), this is $\rx \sim p(\rx)$ with
\begin{equation}
 p(\rx) = \sum_{k=1}^{K} w_k ~ \mathcal{N}(\mu_k, \Sigma_k) ~~.
\end{equation}
Any linear transformation $\rz = \ra + B\rx$ of $\rx \sim p(\rx)$ with $\ra \in \mathbb{R}^n$ and B being an invertible $n \times n$ matrix follows a Gaussian Mixture Model with density function
\begin{equation}
 p(\rz) = \sum_{k=1}^{K} w_k ~ \mathcal{N}(a + B\mu_k, B \Sigma_k B^\top) ~~.
\end{equation}
\end{theorem}

\begin{refproof}[Proof of Theorem \ref{theorem:gmm_lin_trans_appendix}]
With $\rx \in \mathbb{R}^n$ following a Gaussian Mixture Model, its probability density function can be written as
\begin{equation}
 p(\rx) = \sum_{k=1}^{K} w_k ~ \mathcal{N}(\mu_k, \Sigma_k) 
 = \frac{1}{(2 \pi)^{\frac{1}{2}}}  \sum_{k=1}^{K} w_k \frac{\exp \left( -\frac{1}{2} (\rx-\mu_k)^\top \Sigma_k^{-1} (\rx-\mu_k) \right)} {|\Sigma_k|^{\frac{1}{2}}}
\end{equation}
Let $\rz \sim q(\rz)$ be a linear transformation $\rz = \ra + B\rx$ of $\rx \sim p(\rx)$ with $a \in \mathbb{R}^n$ and B being an invertible $n \times n$ matrix. From Lemma \ref{lemma:pdf_lin_trans_appendix} follows that
\begin{align}
 p(\rz) &= \frac{1}{(2 \pi)^{\frac{1}{2}}}  \sum_{k=1}^{K} w_k \frac{\exp \left( -\frac{1}{2} (B^{-1}\rz - B^{-1}\ra -\mu_k)^\top \Sigma_k^{-1} (B^{-1}\rz - B^{-1}\ra-\mu_k) \right)} {|B| ~ |\Sigma_k|^{\frac{1}{2}}} \\
  &= \frac{1}{(2 \pi)^{\frac{1}{2}}}  \sum_{k=1}^{K} w_k 
  \frac{\exp \left( -\frac{1}{2} (\rz - (\ra+B\mu_k))^\top (B^{-1})^\top \Sigma_k^{-1} B^{-1} (\rz - (\ra+B\mu_k)) \right)} {|B|~|\Sigma_k|^{\frac{1}{2}}} \\
  &= \frac{1}{(2 \pi)^{\frac{1}{2}}}  \sum_{k=1}^{K} w_k 
  \frac{\exp \left( -\frac{1}{2} (\rz - (\ra+B\mu_k))^\top (B \Sigma_k B^\top)^{-1} (\rz - (\ra+B\mu_k)) \right)} {|B \Sigma_k B^\top |^{\frac{1}{2}}} \\
    &= \sum_{k=1}^{K} w_k ~ \mathcal{N}(\ra+B\mu_k, B \Sigma_k B^\top)  \tag*{\qed}
\end{align}
\end{refproof}

\subsection{Density Simulation} \label{appendix:density_sim}

This sections holds detailed descriptions of the simulated conditional densities used for the experiments in chapter \ref{sec:sim_evaluation}. The respective conditional densities are illustrated in Figure \ref{fig:sim_densities}.

\subsubsection{EconDensity}
This simple, economically inspired, distribution has the following data generating process $(x,y) \sim p(x,y)$:
\begin{align}
        x &= \left|\epsilon_x\right|, ~~~ \epsilon_x \sim N(0,1) \\
        \sigma_y &= 1 + x \\
        y &= x^2 + \epsilon_y, ~~~ \epsilon_y \sim N(0,\sigma_y)
\end{align}
The conditional density follows as
\begin{equation}
    p(y|x) = \mathcal{N}(y | \mu=x^2, \sigma = 1+x)
\end{equation}
and is illustrated in Figure \ref{fig:econ_density}. One can imagine $x$ to represent financial market volatility, which is always positive with rare large realizations. $y$ can be an arbitrary variable that is explained by volatility. We choose a non-linear relationship between $x$ and $y$ to check how the estimators can cope with that. To make things more difficult, the relationship between $x$ and $y$ becomes more blurry at high $x$ realizations, as expressed in a heteroscedastic $\sigma_y$, that is rising with $x$. This reflects the common behaviour of higher noise in the estimators in times of high volatility.

\subsubsection{ARMAJump}
The underlying data generating process for this simulator is an AR(1) model with a jump component. A new realization $x_t$ of the time-series can be described as follows:
\begin{equation}
x_t = \left[ c(1-\alpha) + \alpha x_{t-1} \right] + (1-z_t) \sigma \epsilon_t + z_t \left[ -c + 3\sigma \epsilon_t \right] \qquad  \epsilon_t \sim N(0,1), z_t \sim B(1,p)
\end{equation}
%1
In that, $c \in \mathbb{R}$ is the long run mean of the AR(1) process and $\alpha \in \mathbb{R}$ constitutes the autoregressive factor, describing how fast the AR(1) time series returns to its long run mean $c$.  Typically an ARMA process is perturbed by Gaussian White Noise $\sigma \epsilon_t$ with standard deviation $\sigma \in \mathbb{R}^+$. We add a jump component, that occurs with probability $p$ and is indicated by the Bernoulli distributed binary variable $z_t$. If a jump occurs, a negative shock of the same magnitude as $c$ is accompanied by Gaussian noise with three times higher standard deviation than normal. The dynamic is a discrete version of the class of affine jump diffusion models, which are heavily used in bond and option pricing.
Here, for each time period $t$, the conditional density $p(x_t|x_{t-1})$ shall be predicted. Note that in this case $y$ corresponds to $x_t$.
The conditional density follows as mixture of two Gaussians:
\begin{equation}
p(x_t|x_{t-1}) = (1-p) \mathcal{N}(x_t| \mu=  c(1-\alpha) + \alpha x_{t-1}, \sigma) + p \mathcal{N}(x_t| \mu=  \alpha (x_{t-1}-c), 3\sigma)
\end{equation}

\begin{figure}
\centering
\begin{subfigure}[b]{0.42\textwidth}
\includegraphics[width=\linewidth]{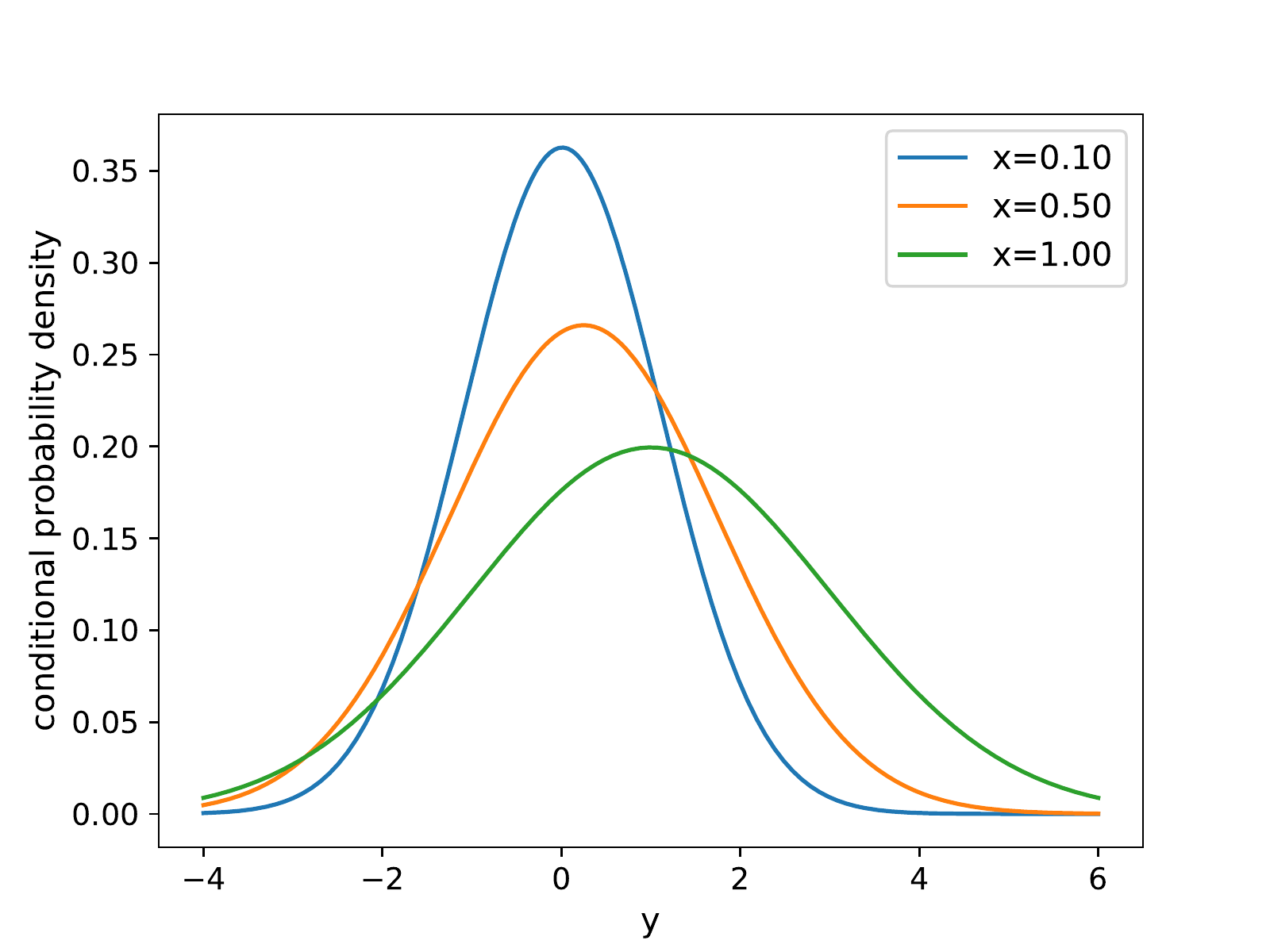}
\caption{EconDensity} \label{fig:econ_density}
\end{subfigure}
\qquad
\begin{subfigure}[b]{0.42\textwidth}
\includegraphics[width=\linewidth]{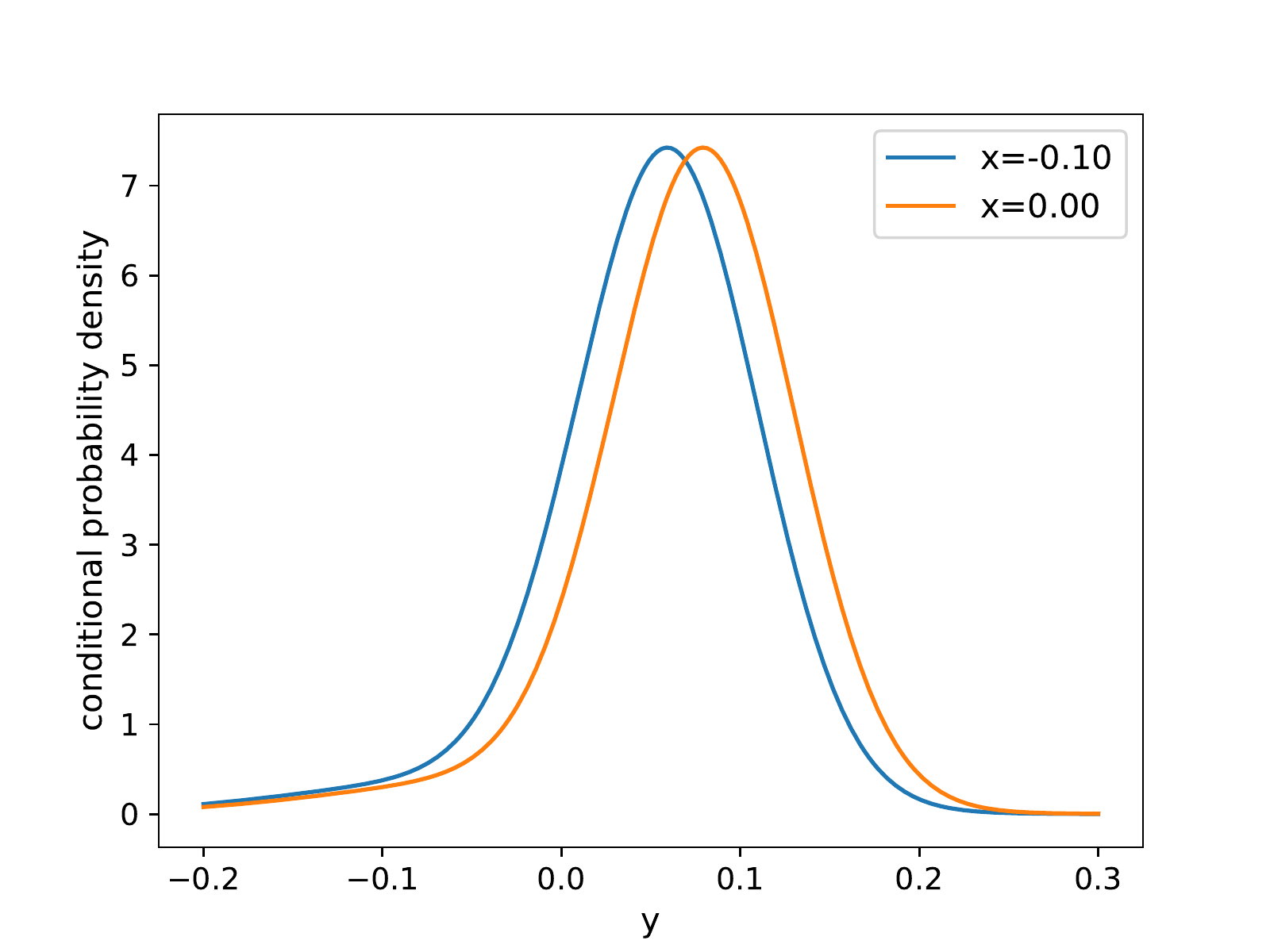}
\caption{ARMAJump} \label{fig:arma_jump}
\end{subfigure}

\begin{subfigure}[b]{0.42\textwidth}
\includegraphics[width=\linewidth]{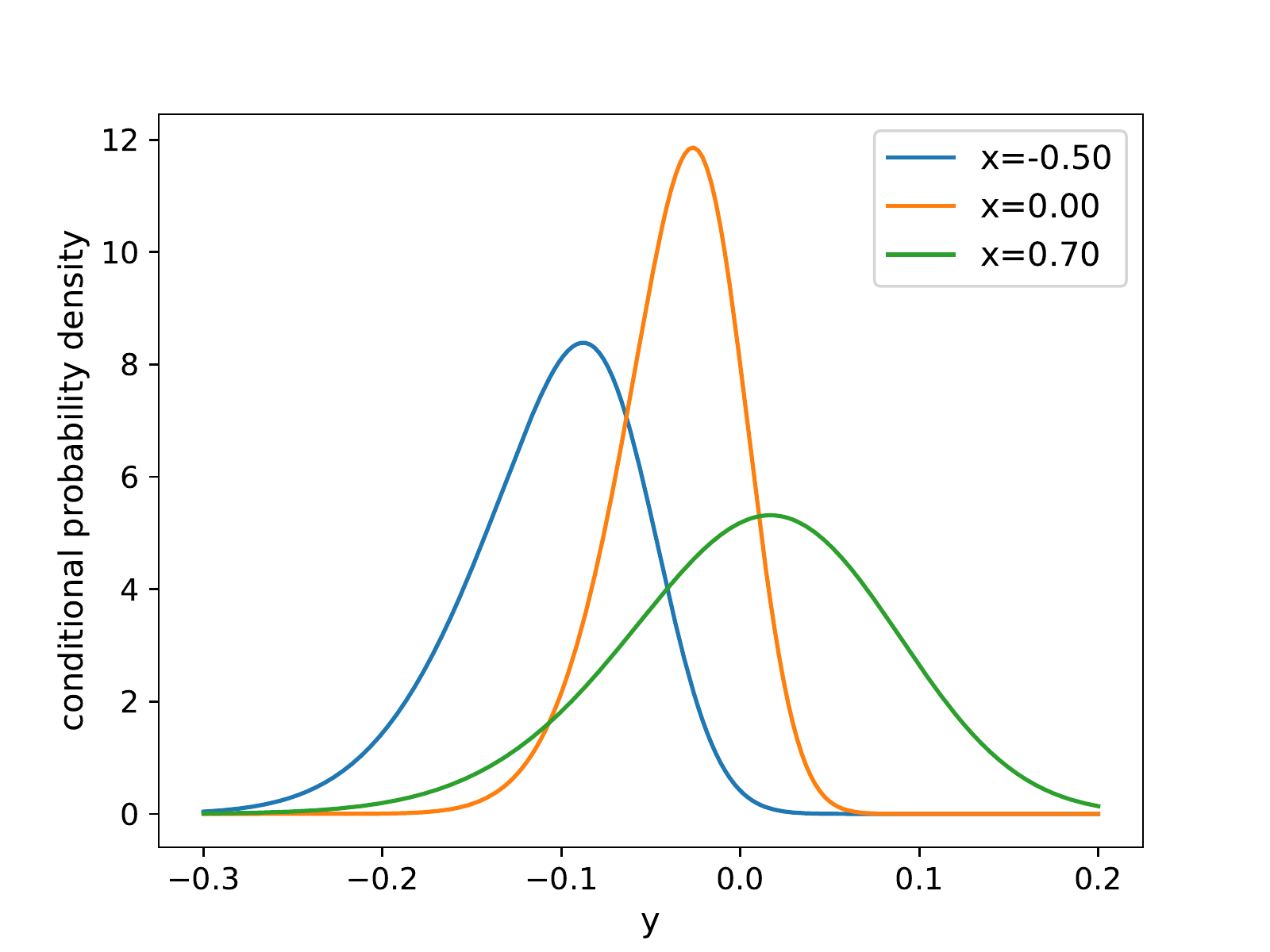}
\caption{SkewNormal} \label{fig:skew_norm}
\end{subfigure}
\qquad
\begin{subfigure}[b]{0.42\textwidth}
\includegraphics[width=\linewidth]{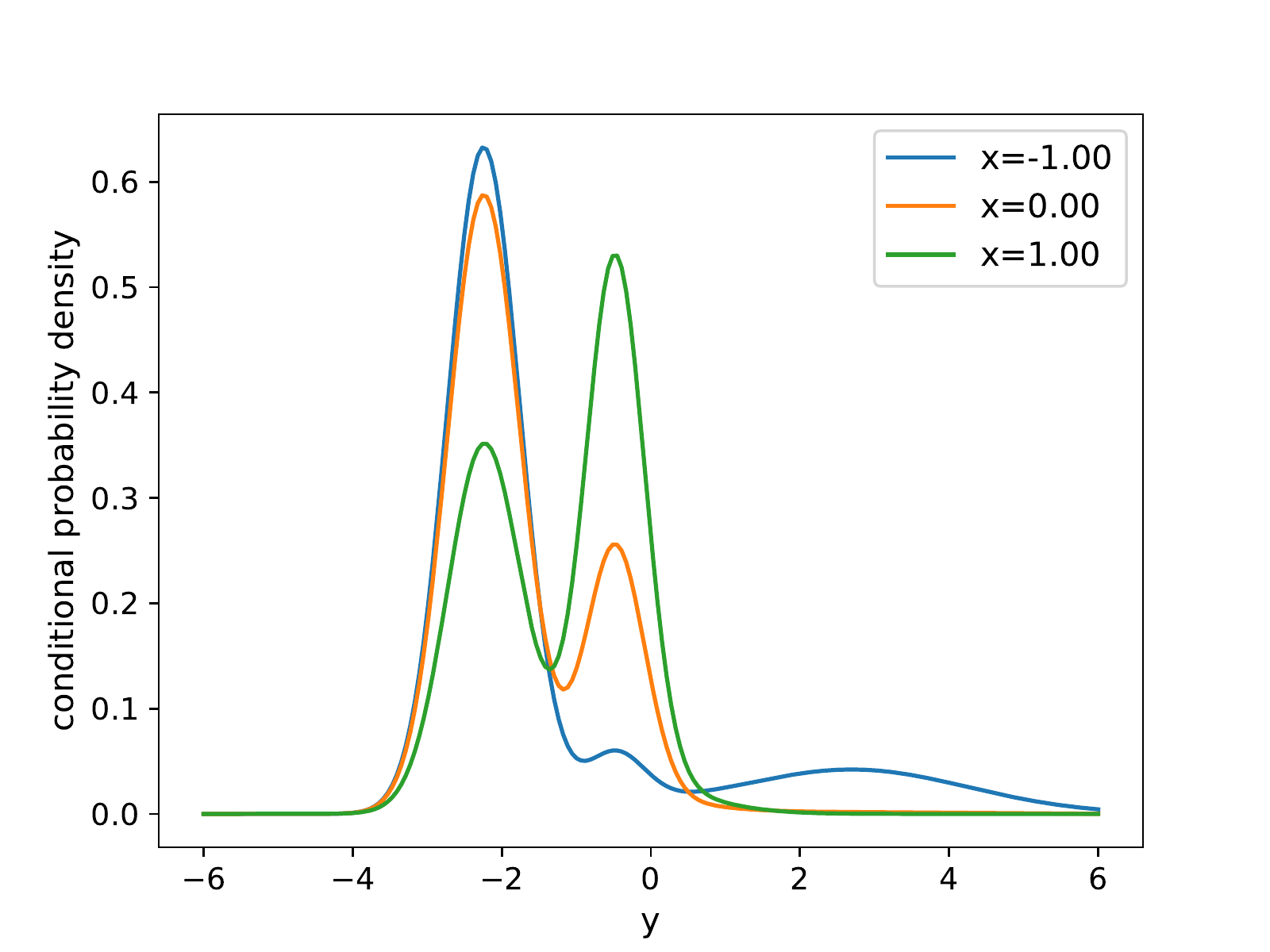}
\caption{GaussianMixture} \label{fig:gmm}
\end{subfigure}

\caption{\textbf{Conditional density simulation models.} Conditional probability densities corresponding to the different simulation models. The coloured graphs represent the probability densities $p(y|x)$, conditioned on different values of $x$.}
\label{fig:sim_densities}
\end{figure}

Figure \ref{fig:arma_jump} depicts the ARMAJump conditional probability density for the time-series parameters $c=0.1, \alpha=0.2, p=0.1, \sigma=0.05$. As can be seen in the depiction, the conditional distribution has a negative skewness, resulting from the jump component.

\subsubsection{SkewNormal}
The data generating process $(x,y) \sim p(x,y)$ resembles a bivariate joint-distribution, wherein $x \in \mathbb{R}$ follows a normal distribution and $y\in \mathbb{R}$ a conditional skew-normal distribution \citep{Andel1984}. The parameters $(\xi, \omega, \alpha)$ of the skew normal distribution are functionally dependent on $x$. Specifically, the functional dependencies are the following:
\begin{align}
        x &\sim \mathcal{N} \left(~ \cdot ~ \bigg \vert \mu=0, \sigma=\frac{1}{2} \right) \\
        \xi(x) &= a*x+b \qquad a, b \in \mathbb{R}\\
        \omega(x) &= c*x^2+d \qquad c, d \in \mathbb{R} \\
        \alpha(x) &= \alpha_{low} + \frac{1}{1+e^{-x}} * (\alpha_{high}-\alpha_{low})  \\
        y &\sim SkewNormal \big( \xi(x), \omega(x), \alpha(x) \big)
\end{align}
Accordingly, the conditional probability density $p(y|x)$ corresponds to the skew normal density function:
\begin{equation}
    p(y|x) = \frac{2}{\omega(x)} \mathcal{N}\left(\frac{y - \xi(x)}{\omega(x)}\right) \Phi \left(\alpha(x)\frac{y - \xi(x)}{\omega(x)}\right)
\end{equation}
In that, $\mathcal{N}(\cdot)$ denotes the density, and $\Phi(\cdot)$ the cumulative distribution function of the standard normal distribution. The shape parameter $\alpha(x)$ controls the skewness and kurtosis of the distribution. We set $\alpha_{low}=-4$ and $\alpha_{high}=0$, giving $p(y|x)$ a negative skewness that decreases as $x$ increases. This distribution will allow us to evaluate the performance of the density estimators in presence of skewness, a phenomenon that we often observe in financial market variables.  Figure \ref{fig:skew_norm} illustrates the conditional skew normal distribution.

\subsubsection{GaussianMixture}
The joint distribution $p(\rx,\ry)$ follows a GMM. We assume that $\rx \in \mathbb{R}^m$ and $\ry \in \mathbb{R}^l$ can be factorized, i.e.
\begin{equation}
    p(\rx,\ry) = \sum_{i=1}^K w_k ~\mathcal{N}(\ry|\mu_{\ry,k}, \Sigma_{\ry,k}) \mathcal{N}(\rx|\mu_{\rx,k}, \Sigma_{\rx,k}) \label{eg:gmm_factorized}
\end{equation}
When $\rx$ and $\ry$ can be factorized as in (\ref{eg:gmm_factorized}), the conditional density $p(\ry|\rx)$ can be expressed as:
\begin{align}
p(\ry|\rx) = \sum_{i=1}^K W_k(x) ~\mathcal{N}(\ry|\mu_{\ry,k}, \Sigma_{\ry,k})
\end{align}
wherein the mixture weights are a function of $\rx$:
\begin{equation}
W_k(x) = \frac{w_k ~ \mathcal{N}(\rx|\mu_{\rx,k}, \Sigma_{\rx,k}) }{\sum_{j=1}^K w_k ~ \mathcal{N}(\rx|\mu_{\rx,j}, \Sigma_{\rx,j})}
\end{equation}
For details and derivations we refer the interested reader to \citet{Sung2004} and \citet{Gilardi2002}.
Figure \ref{fig:gmm} depicts the conditional density of a GMM with 5 components (i.e. $K=5$ and 1-dimensional $\rx$ and $\ry$ (i.e. $l=m=1$).

%%%%%%%%%%%%%%%%%%%%%%%%%%%%%%%%%%%%%%%%%%%%%%%%%%%%%%%%%%%%%%%%%%%%%%%%%%%

\subsection{Conditional Density Estimators in the Benchmark Study}
This section holds a detailed description of the CDE methods, compared to, in the benchmark studies: 
\begin{itemize}

%%%%%%%%%%%%%%%%%%
\item \textbf{Conditional Kernel Density Estimation (CKDE):} This non-parametric conditional density approach estimates both the joint probability $\hat{p}(\rx, \ry)$ and the marginal probability $\hat{p}(\rx)$ with KDE (see Section \ref{sec:nonparametric}). The conditional density estimate follows as density ratio $\hat{p}(\ry|\rx) = \frac{\hat{p}(\rx, \ry)}{\hat{p}(\rx)}$. For selecting the bandwidths $h_\rx$ and $h_\ry$ of the kernels, the rule-of-thumb of \citet{Silverman1982} is employed:
\begin{equation}
    h = 1.06 \hat{\sigma} N^{-\frac{1}{4 + d}} \label{eq:rule_of_thumb}
\end{equation}
In that, $N$ denotes the number of samples, $\hat{\sigma}$ the empirical standard deviation and $d$ the dimensionality of the data. The rule-of-thumb assumes that the data follows a normal distribution. If this assumption holds, the selected bandwidth $h$ is proven to be optimal w.r.t. the \textit{IMSE} criterion.

%%%%%%%%%%%%%%%%%
\item \textbf{CKDE with bandwidth selection via cross-validation (CKDE-CV):} Similar to the CDKE above but the bandwidth parameters $h_\rx$ and $h_\ry$ are determined with leave-one-out maximum likelihood cross-validation. See \citet{Li2007} for further details about the cross-validation-based bandwidth selection.

%%%%%%%%%%%%%%%%%%%%
\item \textbf{$\epsilon$-Neighborhood kernel density estimation (NKDE):} 
For estimating the conditional density $p(\ry|\rx)$, $\epsilon$-neighbor kernel density estimation employs standard kernel density estimation in a local $\epsilon$-neighborhood around a query point $(\rx, \ry)$ \citep{Sugiyama2010}.

NKDE is similar to CKDE, as it uses kernels, placed in the training data points, to estimate the conditional probability density. However, rather than estimating both the joint probability $p(\rx,\ry)$ and marginal probability $p(\rx)$, NKDE forms a density estimate by only considering a local subset of the training samples $\{(\rx_i, \ry_i)\}_{i \in \mathcal{I}_{\rx, \epsilon}}$, where $\mathcal{I}_{\rx, \epsilon}$ is the set
of sample indices such that $||\rx_i - \rx||_2 \leq \epsilon$.
The estimated density can be expressed as

\begin{equation}
p(\ry|\rx) = \sum_{j \in \mathcal{I}_{x, \epsilon}} w_j ~ \prod_{i=1}^l \frac{1}{h^{(i)}} K \left(\frac{y^{(i)}~ - y_j^{(i)}}{h^{(i)}} \right)
\end{equation}

wherein $w_j$ is the weighting of the j-th kernel and $K(z)$ a kernel function. In our implementation $K$ is the density function of a standard normal distribution. The weights $w_j$ can either be uniform, i.e. $w_j = \frac{1}{|\mathcal{I}_{x, \epsilon}|}$ or proportional to the distance $||x_j - x||$. The vector of bandwidths $\rh=(h^{(1}, ..., h^{(l)})^T$ is determined with the rule-of-thumb (see Equation \ref{eq:rule_of_thumb}), where the the number of samples $N$ corresponds to the average number of neighbors in the training data:
\begin{equation}
    N =  \frac{1}{N} \sum_{n=1}^N |\mathcal{I}_{\rx_n, \epsilon}| - 1
\end{equation}

%%%%%%%%%%%%%%%%%%
\item \textbf{Least-Squares Conditional Density Estimation (LSCDE):} A semi-parametric estimator that computes the conditional density as linear combination of kernels \citep{Sugiyama2010}.
\begin{equation}
\hat{p}_{\alpha}(\ry|\rx) \propto \alpha^T \phi(\rx,\ry)
\end{equation}
Due to its restriction to linear combinations of Gaussian kernel functions $\phi$, the optimal parameters $\alpha$ w.r.t. the integrated mean squared objective
\begin{equation}
J(\alpha) = \int \int ( \hat{p}_{\alpha}(\rx,\ry) - p(\rx,\ry))^2 p(\rx) d\rx d\ry
\end{equation}
objective can be computed in closed form. However, at the same time, the linearity assumption makes the estimator less expressive than the KMN or MDN. See Appendix \citet{Sugiyama2010} for details.
\end{itemize}

%%%%%%%%%%%%%%%%%%%%%%%%%%%%%%%%%%%%%%%%%%%%%%%%%%%%%%%%%%%%%%%%%%%%%%%%%5
\subsection{Hyper-Parameter Settings and Selection}
\label{appendix:hyperparam}
This sections lists and describes the hyper-parameter configurations used for the empirical evaluations. In that, we first attend to the default configuration used in the benchmarks in Fig. \ref{fig:sim_benchmark} and Table \ref{table:eurostoxx_eval}. 

The neural network has two hidden layers with 16 neurons each, tanh non-linearities and weight normalization \citep{Salimans2016b}. For the KMN, we use $K=50$ Gaussian mixture components and, for the MDN, $K=20$ components. The neural network is trained for 1000 epochs with the Adam optimizer \citep{Kingma2015}. In that, the Adam learning rate is set to $\alpha=0.001$ and the mini-batch size is 200. 
In order to select the kernel centers, for the KMN, we employ K-means clustering on the $\ry_i$ data points. In each of the selected centers, we place two Gaussians with initial standard deviation $\sigma_1=0.7$ and $\sigma_2=0.3$. While the locations of the Gaussians are fixed during training, the scale / standard deviation parameters are trainable and thus adjusted by the optimizer. Table \ref{tab:standard_params} provides an overview of the default hyper-parameters.

\begin{table}[h]
    \centering
    \begin{tabular}{|l|c|c|}
        \hline
        & MDN & KMN \\ \hline
        hidden layer sizes & (16,16) & (16,16)  \\ \hline
        hidden non-linearity & tanh  & tanh \\ \hline
        training epochs & 1000  & 1000 \\ \hline
        Adam learning rate  & 0.001 & 0.001 \\ \hline
        $K$: number of components & 20  & 50 \\ \hline
        $\eta_x$: noise std x & 0.2  & 0.2 \\ \hline
        $\eta_y$: noise std y & 0.1  & 0.1 \\ \hline
        weight normalization & True  & True \\ \hline
        data normalization & True  & True \\ \hline
        initialization of scales & -  & [0.7, 0.3] \\ \hline
        trainable scales & -  & True \\ \hline
    \end{tabular}
    \caption{$\quad$ Default hyper-parameter configuration for MDN and KMN}
    \label{tab:standard_params}
\end{table}

In the benchmark study with hyper-parameter selection (see Fig. \ref{table:eurostoxx_eval_cv}), a subset of the parameters has been optimized via 10-fold cross-validation grid search. Since the search space grows exponentially with the number of parameters to optimize over, we constrained the grid search to the hyper-parameters which we found to have the greatest influence on the estimators performance. In the case of MDN and KMN, these are the number of training epochs, the number of mixture components and the noise regularization intensities. For LSCDE, the search comprised the number of kernels, bandwidth and damping parameter $\lambda$. Table \ref{tab:cv_params} holds the hyper-parameters that were determined through the parameter search and subsequently used to fit and evaluate the respective estimator.

\begin{table}[h]
    \centering
    \begin{tabular}{|r|c|c|c|}
        \hline
        & MDN-CV & KMN-CV & LSCDE-CV\\ \hline
        training epochs & 500  & 500 & -\\ \hline
        $K$: number of components & 10  & 200 & 1000\\ \hline
        $\eta_x$: noise std x & 0.3  & 0.2 & -\\ \hline
        $\eta_y$: noise std y & 0.15  & 0.15 & - \\ \hline
        bandwidth & - & - & 0.5 \\ \hline
        $\lambda$: LSCDE damping parameter & - & - & 0.1  \\ \hline
    \end{tabular}
    \caption{$\quad$ Hyper-parameter configuration determined with 10-fold cross-validation on the EuroStoxx 50 data set}
    \label{tab:cv_params}
\end{table}

%%%%%%%%%%%%%%%%%%%%%%%%%%%%%%%%%%%%%%%%%%%%%%%%%%%%%%%%%%%%%%%%%%%%%%%%%%%%

\subsection{The Euro Stoxx 50 Data} \label{appendix:eurostoxx_data}
The following section holds a detailed description of the EuroStoxx 50 dataset employed in chapter \ref{sec:eurostoxx_eval}. The data comprises 3169 trading days, dated from January 2003 until June 2015. We define the task as predicting the conditional probability density of 1-day log returns, conditioned on 14 explanatory variables. These conditional variables are listed below: 

\begin{itemize}
     \item \textbf{log\_ret\_last\_period:} realized log-return of previous trading day\\
     \item \textbf{log\_ret\_risk\_free\_1d:} risk-free 1-day log return, computed based on the overnight index swap rate (OIS) with 1 day maturity. The OIS rate $r_f$ is transformed as $log(\frac{r_f}{365}+1)$.\\
     \item \textbf{RealizedVariation:} estimate of realized variance of previous day, computed as sum of squared 10 minute returns over the previous trading day \\
     \item \textbf{SVIX:} 30-day option implied volatility\footnotemark[3] \citep{Whaley1993} \\
     \item \textbf{bakshiKurt:} 30-day option implied kurtosis\footnotemark[3] \citep{Bakshi2003} \\
     \item \textbf{bakshiSkew:} 30-day option implied skewness\footnotemark[3] \citep{Bakshi2003} \\
     \item \textbf{Mkt-RF:} Fama-French market return factor \citep{Fama1993} \\
     \item \textbf{SMB:} Fama-French Small-Minus-Big factor \citep{Fama1993}\\
     \item \textbf{HML:} Fama-French High-Minus-Low factor \citep{Fama1993}\\
     \item \textbf{WML:} Winner-Minus-Looser (momentum) factor \citep{Carhart1997} \\
     \item \textbf{Mkt-RF 10-day risk:} risk of market return factor; sum of squared market returns over the last 10 trading days\\
     \item \textbf{SMB 10-day risk:} SMB factor risk; sum of squared factor returns over the last 10 days\\
     \item \textbf{HML 10-day risk:} HML factor risk; sum of squared factor returns over the last 10 days\\
     \item \textbf{WML 10-day risk:} WML factor risk; sum of squared factor returns over the last 10 days
\end{itemize}

\footnotetext[3]{The option implied moments are computed based on a options with maturity in 30 days. Since the days to maturity vary, linear interpolation of the option implied moments, corresponding to different numbers of days till maturity, is used to compute an estimate for maturity in 30 days.}

%%%%%%%%%%%%%%%%%%%%%%%%%%%%%%%%%%%%%%%%%%%%%%%%%%%%%%%%%%%%%%%%%%%%%%%%%%%%

\subsection{Estimation of the Conditional Moments} 
\label{appendix:compute_moments}
This section briefly describes how the moments of the conditional density estimates $\hat{p}(\ry|\rx)$ are computed. In particular, we will focus on the mean, covariance, skewness and kurtosis. 

By default the centered moments are estimated via numerical or monte-carlo integration, using their respective definitions:
\begin{align}
    \hat{\mu}(\rx) &= \int_\mathcal{Y} \ry \hat{p}(\ry|\rx) d\ry \\
    \hat{\sigma}(\rx) &= \sqrt{\int_\mathcal{Y} (\ry - \hat{\mu}(\rx))^2 \hat{p}(\ry|\rx) d\ry} \\
    \widehat{Cov}(\rx) &= \int_\mathcal{Y} (\ry - \hat{\mu}(\rx)) (\ry - \hat{\mu}(\rx))^T \hat{p}(\ry|\rx) d\ry \\
     \widehat{Skew}(\rx) &= \int_\mathcal{Y} \left(\frac{\ry - \hat{\mu}(\rx)}{\hat{\sigma}(\rx)}\right)^3 \hat{p}(y|\rx) dy \\
      \widehat{Kurt}(\rx) &= \int_\mathcal{Y} \left(\frac{y - \hat{\mu}(\rx)}{\hat{\sigma}(\rx)}\right)^4 \hat{p}(y|\rx) dy - 3
\end{align}
Our implementation only supports estimating skewness and kurtosis for univariate target variables, i.e. $dim(\mathcal{Y})=1$. If $dim(\mathcal{Y})=1$, the integral is approximated with numerical integration, using the Gaussian quadrature with 10000 reference points, for which the density values are calculated. If $dim(\mathcal{Y}) > 1$, we use Monte-Carlo integration with 100,000 samples (see Appendix\ref{appendix:monte_carlo}).

In case of the KMN and MDN, the conditional distribution is a GMM. Thus, we can directly calculate mean and covariance from the GMM parameters, outputted by the neural network. The mean follows straightforward as weighted sum of the Gaussian component centers: $\mu_{k}(\rx;\theta)$
\begin{equation}
    \hat{\mu}(\rx) = \sum_{k=1}^K w_{k}(\rx;\theta) \mu_{k}(\rx;\theta)
\end{equation}
The covariance matrix can be computed as
\begin{equation}
\hat{Cov}(\rx) = \sum_{k=1}^K w_{k}(\rx;\theta) \left( (\mu_{k}(\rx;\theta) - \hat{\mu}(\rx)) (\mu_{k}(\rx;\theta) - \hat{\mu}(\rx))^T + diag(\sigma_{k}(\rx;\theta)^2)\right)
\end{equation}
wherein the outer product accounts for the covariance that arises from the different locations of the components and the diagonal matrix for the inherent variance of each Gaussian component.

\clearpage

% Bibliography.

\begin{doublespacing}   % Double-space the bibliography
\bibliographystyle{jf}
\bibliography{references}
\end{doublespacing}

\clearpage

% Print end notes
\renewcommand{\enotesize}{\normalsize}
%\begin{doublespacing}
%  \theendnotes
%\end{doublespacing}

% Figures and tables, showing how to structure captions
\clearpage
\end{document}